\documentclass[10pt,journal,compsoc]{IEEEtran}

\usepackage{times}
\usepackage{epsfig}
\usepackage{graphicx}
\usepackage{amsmath}
\usepackage{amssymb}
\usepackage{subfigure}
\usepackage{amsthm}
\usepackage{algorithm,algpseudocode}
\usepackage{adjustbox}
\usepackage[dvipsnames]{xcolor}
\usepackage{pythonhighlight}

\usepackage[pagebackref=true,breaklinks=true,letterpaper=true,colorlinks,bookmarks=false]{hyperref}
\usepackage{multicol}
\usepackage{multirow}
\newcommand{\tablestyle}[2]{\setlength{\tabcolsep}{#1}\renewcommand{\arraystretch}{#2}\centering\footnotesize}
\newlength\savewidth\newcommand\shline{\noalign{\global\savewidth\arrayrulewidth
  \global\arrayrulewidth 1pt}\hline\noalign{\global\arrayrulewidth\savewidth}}

\newcommand{\apbbox}[1]{AP$^\text{bb}_\text{#1}$}
\newcommand{\apmask}[1]{AP$^\text{mk}_\text{#1}$}
\newcommand{\apkp}[1]{AP$^\text{kp}_\text{#1}$}

\input{def.set}
\usepackage{listings}
\ifCLASSOPTIONcompsoc
  \usepackage[nocompress]{cite}
\else
  \usepackage{cite}
\fi

\ifCLASSINFOpdf

\else

\fi

\begin{document}
%
\title{A Low Rank Promoting Prior\\ for Unsupervised Contrastive Learning}

\author{Yu~Wang, Jingyang~Lin, Qi Cai, Yingwei~Pan,\\ Ting~Yao,~\IEEEmembership{Member, IEEE}, Hongyang Chao,~\IEEEmembership{Member, IEEE,} Tao~Mei,~\IEEEmembership{Fellow,~IEEE}
\IEEEcompsocitemizethanks{\IEEEcompsocthanksitem Y. Wang, Y. Pan, T. Yao, T. Mei are with JD AI Research, China
\IEEEcompsocthanksitem J. Lin and H. Chao are with SUN YAT-SEN University, China
\IEEEcompsocthanksitem Q. Cai is with University of Science and Technology of China, China
\IEEEcompsocthanksitem Email to Yu Wang: feather1014@gmail.com}
\thanks{This work has been submitted to the IEEE for possible publication. Copyright may be transferred without notice, after which this version may no longer be accessible.}

}


\IEEEtitleabstractindextext{%
\begin{abstract}
 Unsupervised learning is just at a tipping point where it could really take off. Among these approaches, contrastive learning has seen tremendous progress and led to state-of-the-art performance. In this paper, we construct a novel probabilistic graphical model that effectively incorporates the low rank promoting prior into the framework of contrastive learning, referred to as LORAC. In contrast to the existing conventional self-supervised approaches that only considers independent learning, our hypothesis explicitly requires that all the samples belonging to the same instance class lie on the same subspace with small dimension. This heuristic poses particular joint learning constraints to reduce the degree of freedom of the problem during the search of the optimal network parameterization. Most importantly, we argue that the low rank prior employed here is not unique, and many different priors can be invoked in a similar probabilistic way, corresponding to different hypotheses about underlying truth behind the contrastive features. Empirical evidences show that the proposed algorithm clearly surpasses the state-of-the-art approaches on multiple benchmarks, including image classification, object detection, instance segmentation and keypoint detection.
\end{abstract}

}

\maketitle

\IEEEdisplaynontitleabstractindextext

\IEEEpeerreviewmaketitle

\IEEEraisesectionheading{\section{Introduction}\label{sec:introduction}}

\IEEEPARstart{U}{nsupervised} learning has brought a powerful and successful revolution in regime of feature learning. One primary incentive is that the underlying high dimensional feature behind the data intrinsically has a much richer pattern than what the human annotation defines. By relying merely on the data itself, unsupervised learning sees beyond labels and seeks to characterize the underlying feature distributions completely out of the constraints of human crafted annotation. Contrastive learning is a pioneering class of unsupervised algorithms that receives much attention nowadays. In short, the central goal of contrastive learning is to train a machine to classify between ``similarity'' and ``distinctiveness'', such that the model parameters favor invariant features among similar instances and promote discriminative features between distinct instances. Recent research demonstrates that such instance-level in-class invariance is beneficial when deploying further downstream tasks on the features pre-trained under such contrastive learning assumption \cite{chen2020simple, he2019momentum,oord2018cpc,tian2019contrastive, Jingpami2020}.

While self-supervised learning approaches display impressive performance when employed as efficient pre-training techniques, existing methods nevertheless ignored useful dependency among multiple samples. We therefore take a non-trivial step further by probing alternative hypotheses under the contrastive learning framework. In detail, we leverage on the assumption that all samples within the same instance class lie near some low-dimensional subspace, i.e., when stacking all the feature points belonging to the same instance class as row vectors of a matrix, the matrix should be low rank {\cite{zhang2017split}}. This assumption is motivated by two important observations. Firstly, as the contrastive learning procedure proceeds, the representations of positive pairs are intrinsically encouraged to be close.  Secondly, regardless of contrastive learning, low rank assumption has been shown appealing in many computer vision tasks. These assumptions roots in the fact that many image and video data have low intrinsic dimensionality, lying on low-dimensional subspace. Take for instance, the face images from the same person under different shadowing conditions or brightness have shown to be lying on the same low-dimensional subspace \cite{elhamifar2013sparse}; Frames from a slowly evolving sequence of surveillance video data usually can be modeled as a composition of low rank matrix plus an additional sparse innovation matrix \cite{MaYilowrank2013}; Motion trajectories belonging to the same object in a video sequence come from the same subspace \cite{kanatani2001motion}; Many other applications include \cite{candes2011robust, liu2011latent, MaLlowrank2012, Zhang2013lowrank}.

In regime of contrastive learning, refraining the positive pairs to live within the same class simultaneously encourage multiple views to respect a joint constraint. Intuitively, this constraint reduces the degree of the freedom of the optimization problem and avoid bad local minima. Given this motivation, we propose a novel probabilistic model that involves a LOw RAnk promoting prior for Contrastive learning, named ``LORAC''. The proposed method are motivated from a probabilistic view of point, while it can be conveniently implemented via conventional contrastive learning architectures \cite{he2019momentum} with closed form of loss function.

Our contributions in this paper include: {\bf 1.} We construct a novel probabilistic model ``LORAC'' that effectively incorporates the low rank promoting prior into the framework of contrastive learning. We show that many different probabilistic priors can be invoked to serve similar purpose, corresponding to different hypotheses about underlying truth behind the unsupervised learning model. {\bf 2.} By relaxing the low rank approximation via nuclear-norm minimization, we obtain an analytic form of loss that allows for end-to-end training of the deep network. The loss is shown to be effectively shrinking the nuclear norm of the constructed matrix, therefore constraining the total subspace dimension that all augmentations of an image can live in. {\bf 3.} We theoretically discuss the mathematical interpretations behind the loss, showing that under certain form of the prior construction, LORAC elusively removes bad local minima, without increasing the conventional contrastive loss.

\vspace{-0.4cm}
\section{Related Work}
{\bf{Self-supervised Learning}}.
Self-supervised learning (SSL), as its name implies, waives the need of human annotation during training. The motivation behind is to best leverage the enormous unlabeled data in an economic way, while unlabeled data intrinsically has richer structure than what the sparse labels describes and is worthy of further exploration. Without access to manual labels, SSL methods usually resort to the predefined ``pretext tasks'' to gain some form of supervision from the data itself. The resultant feature extractor out of such self-supervised learning procedure then is expected to provide a good initialization of the network for downstream supervised tasks. Take for instance, in \cite{gidaris2018unsupervised}, the network learns to predict the rotation angle of each image in order to improve the semantics of features through such pretext training tasks. In \cite{doersch2015unsupervised}, the patches from a single image are sampled, and the feature extractor is forced to learn the spatial arrangements of the patches. Similarly, in \cite{noroozi2016unsupervised}, the training objective is to solve jigsaw puzzles for each image, i.e., to predict the permutation of patches composing each training image. Regarding the work in \cite{zhang2017split}, each training data is split into different channels. The network is then trained to reconstruct a subset of the data channels from the remaining split of the seen channels. Representative self-supervised learning approaches that rely on specific pretext tasks also include  \cite{deshpande2015learning,iizuka2016let,larsson2017colorization,zhang2016colorful,zhang2017split,carlucci2019domain, goyal2019scaling,Liupami2019}. Orthogonal to those pretext task based approaches though, contrastive learning methods and the associated extensions have demonstrated strong performance. We briefly discuss such line of methods in the following section.

{\bf{Contrastive Learning and Non-contrastive Learning}}.
Contrastive learning techniques are recently considered as the most competitive self-supervised learning approaches \cite{bachman2019learning,chen2020simple,he2019momentum,henaff2019data,hjelm2018learning,laskin2020curl,misra2019self,oord2018cpc,tian2019contrastive,  wu2018unsupervised,ye2019unsupervised,zhuang2019local}. Inspired from the seminal work in \cite{nce2010noise}, the central goal of these contrastive approaches is to maximize the mutual information of different views of the same images in the latent feature space, while negative samples from other instances remain ``contrastive''. In \cite{he2019momentum, chen2020improved}, the proposed MoCo models conveniently maintain a memory bank so that the contrastive loss can access a large number of negative samples, approaching the theoretical bound proposed in \cite{nce2010noise} and  \cite{oord2018cpc}. In BYOL, the authors rather remove the negative samples from training and introduces two neural networks, referred to
as online and target networks that learn from each other. BYOL does not involve negative samples, in which sense the contrastive flavor is absent. However, BYOL still well inherits the spirit from mainstream contrastive learning approaches that augmented views of the same image should be predicative of each another. The Simsiam approach further explores the role of frozen gradients, momentum encoder, and projection head in the self-supervised learning in \cite{chen2020exploring}. Surprisingly, SimSiam can still well prevent learning collapsing even if the momentum update of the encoder is further removed, as long as the frozen gradient implementation is deployed. Following this line of work, Barlow Twins seeks to directly improve intra-positive pair similarities via the cross-correlation matrix between the views from the same image \cite{barlowtwins2021}. There are also many theoretical works that push forward the understanding of the contrastive learning mechanisms. Arora et al. proves in \cite{saunshi19a} that unsupervised training via contrastive loss can effectively lower the upperbound of the error in downstream supervised tasks, when pre-training data and downstream data come from the same classes.  In \cite{wang20k}, Wang et al. discovers that the effectiveness from contrastive loss boils down to feature alignment and uniformity on the hypersphere. They also show that directly optimizing for these two metrics leads to representations with comparable at downstream tasks. In \cite{tian21a}, by simplifying the deep network into a two-layer linear model, Tian et al. analyze the behavior of non-contrastive SSL, attempting to answer why non-contrastive learning approaches such as BYOL and SimSiam can avoid collapsed representations even if negative samples are absent.  Recently, more and more interesting work extending the idea of contrastive learning and non contrastive learning approaches are emerging \cite{chen2020big,chen2020exploring,chuang2020debiased,foster2020improving,li2020prototypical,mitrovic2020representation,robinson2020contrastive,tian2019contrastive,wu2020conditional,yao2021seco,yu2021selfsupervised}.

{\bf{Multi-View Variants for Contrastive Learning}}.
Many active research focuses on the influence of view selections and multi-view training for contrastive learning \cite{cai2020joint,Swav2020,HardNegMix2020,khosla2020supervised,Miech2020MILNCE,Song2020multicpc,tian2020makes}. These works consistently verify that adding more positives into contrastive learning is beneficial for downstream tasks. Particularly, in \cite{xiao2020should}, the approach generates multiple views and each contrastive loss leaves one type of augmentation out, so that the downstream tasks do not overfit to any specific of view of the training samples. In \cite{tian2019contrastive}, contrastive losses constructed under different views are simply summed together. In \cite{cai2020joint}, the proposed JCL loss even takes into account the effect of infinite number of views under specific generative process definition, which forms an upperbound of their motivating objective. In \cite{Song2020multicpc}, the approach involves multiple positive pairs framed as a multi-label learning problem, aiming at a theoretical tighter bound to query-key mutual information than conventional contrastive learning. \cite{Miech2020MILNCE} propose a loss that is particularly tailored for caption-frames misalignments issues inherent in narrated videos, and can be well generalized to scenarios where noise is inevitable. In SwAV \cite{Swav2020}, the authors introduced the multi-crop augmentation strategy that replaces the conventional two view augmentation with a mix of multiple views of different resolutions, and update online clustering procedure among multiple views. Several works also explore benefits from extra generative augmentations \cite{lee2021imix, tian2020makes} in order to improve existing contrastive learning pipelines. LORAC falls into the category of multi-view contrastive learning, which aims to leverage the dependency between views from the same training image.

{\bf{Comparisons to Existing Work}}. Beyond existing literature, LORAC poses subspace constraint on the contrastive learning approaches {\emph{by first time}}. In this paper, we endeavor to support this rationale both empirically and theoretically. While various interesting view construction strategies are orthogonal and well complementary to our work, we herein investigate a hypothesis of low rank assumption on self-supervised learning. The low rank prior is an alternative assumption that exploits multi-positive pairs under the contrastive learning framework. The low rank hypothesis explicitly requires that positive samples corresponding to the same instance-class should lie on the same low-dimensional subspace. During the training, LORAC searches the least number of eigenvectors that span the whole subspace of augmentation pools, which corresponds to a fundamentally different assumption and associated optimization problem on top or conventional contrastive learning approaches.

\section{Method}\label{sec:CLLR}

In this section, we introduce the low rank hypothesis for contrastive learning. Our motivation is to simultaneously reflect the intrinsic dependencies among multiple views based on low rank assumption. The most compelling part of our proposal is that the incorporation of the low rank prior intrinsically favors correlation among all the positive keys related to the same instance. Specifically, we construct a novel probabilistic graphical model (PGM) rigorously emanated from our hypothesis, followed by the closed form of loss function. In the regime of convex optimization, various surrogate functions have been proposed to handle the low rank assumptions, as the original rank optimization problem is non-convex. Among these options, we employ the simple heuristic where singular value of $\bQ$ upper bounds the rank regularizer \cite{Boyd2001rank}. This surrogate function conveniently allows for end-to-end training of our method.

\subsection{Preliminaries}
{\bf{Contrastive Learning}}. The basic motivation for contrastive learning is to learn features so that samples from different distributions keep far away. We define ``positive keys'' as the samples from the same distribution as a specific ``query'', whereas ``negative keys'' of this query are often assumed from alternative noise distributions. Contrastive loss returns low values if query feature is similar to its positive key feature, while it is distinct from negative keys.

A popular strategy for constructing positive/negative samples is as follows \cite{he2019momentum}. Given a specific training image $\bx_i $ (also referred to as the $i$ instance), we generate a couple of augmented views of $\bx_i $ respectively into the query image $\bx_i^q$, and the positive key image $\bx_i^{k}$. Every $\bx_i^q$ is subsequently mapped into embedding $\bq_i \in {\mathbb R}^d$ via the query encoder $f(\bx_i^{q},  \btheta_q)$, while each $\bx_i^{k}$ is converted to $\bk_{i}^+\in {\mathbb R}^d$ via key encoder $g(\bx_i^{k},\btheta_k)$. In the meanwhile, the negative keys $\bk_{i,j}^{-}$  are generated from some alternative instances distinct from $i$, and form the {\em{noise}} set ${\mathcal N}^{-}=\{\bk_{i,1}^{-},\bk_{i,2}^{-}...\bk_{i,K}^{-}\}$. Both functions $f(\cdot, \btheta_q)$ and $g(\cdot,\btheta_k)$ are implemented using deep neural networks, of which the network parameters $\btheta_q$ and $\btheta_k$ are learned during training.

Noise Contrastive Estimation (NCE) \cite{nce2010noise} verifies that negative samples can be modeled as noises with regard to each query. Take for instance, InfoNCE \cite{oord2018cpc} presents a prevailing contrastive loss based on a softmax formulation:
\begin{equation}
	\label{eq:moco}
	\small
	\mathcal{L}_{CL}=-\frac{1}{N}\sum_{i}^{N} \log \frac{\exp({{\bq_i^T \bk_i^{+}/\tau}})}  {\exp({{\bq_i^T \bk_i^{+}/\tau}})+\sum_{j=1}^{K} \exp({{\bq_i^T \bk_{i,j}^{-}/\tau}})}.
\end{equation}
Eq. (\ref{eq:moco}) explicitly requires the learning to distinguish the positive pairs ($\bq_i $, $\bk_i^{+}$) from the $K$ negative pairs, ($\bq_i $, $\bk_{i,j}^{-}$), $j \in [1,K]$ . Here the hyperparameter  $\tau$ is the predefined temperature following \cite{chen2020simple, he2019momentum, hinton2015distilling}. $N$ is the batchsize.

{\bf Probabilistic Interpretation.} The statistical motivation behind Eq. (\ref{eq:moco}) is simple: given training dataset $X$, we generate $1+K$ number of key features composed of $\bk^+_{i}$ and $\bk^-_{i,j}, j\in [1,K]$. The goal is to maximize the probability that $\bk^+_{i}$ is correctly classified as positive drawn from conditional distribution $p(\bk^+_{i}|\bq_i)$, rather than from the noise distribution $p(\bk^-_{i,j})$ \cite{oord2018cpc}:
\begin{align}
&\small p(v[\bk^+_{i}]=1 |X,\bq_i)= \nonumber\\
&\small \frac{p(\bk^+_{i}|\bq_i)\prod_j p(\bk^-_{i,j})}{p(\bk^+_{i}|\bq_i)\prod_j p(\bk^-_{i,j})+\sum_{j=1}^{K} p(\bk^-_{i,j}|\bq_i)p(\bk^+_i)\prod_{\ell \neq j} p(\bk^-_{i,\ell})}.\label{eq:cpc}
\end{align}

Function $v[\bk^+_{i}]=1$ indicates that $\bk^+_{i}$ is recognized as positive key. By dividing $p(\bk^+_i) \prod_{j=1}^K p(\bk^-_{i,j})$ from both numerator and nominator, Eq. (\ref{eq:cpc}) immediately reduces to:
\begin{align}
&\small p(v[\bk^+_{i}]=1 |X,\bq_i)=\nonumber\\
&\hspace{1.2cm}\small \frac{p(\bk^+_{i}|\bq_i) / p(\bk^+_{i})  }{p(\bk^+_{i}|\bq_i) / p(\bk^+_{i}) +\sum_{j=1}^{K} p(\bk^-_{i,j}|\bq_i) / p(\bk^-_{i,j}) }.\label{eq:cpc1}
\end{align}
By far the most common distributional assumption for conventional contrastive learning is to consider the density ratio as the mutual information via the model \cite{oord2018cpc}:
\begin{align}
\label{eq:density:cpc}
&\small p(\bk^+_{i}|\bq_i) / p(\bk^+_{i}) \propto \exp({{\bq_{i}^T \bk^+_{i}/\tau}}), \\
&\small p(\bk^-_{i,j}|\bq_i) / p(\bk^-_{i,j}) \propto \exp({{\bq_{i}^T \bk^-_{i,j}/\tau}}),
\end{align}
under which we recover the loss Eq. (\ref{eq:moco}). During training, the parameters of the model are optimized so that the chance of correctly classifying $\bk_i^+$ (i.e., Eq. (\ref{eq:cpc})) is maximized.

\textbf{Low Rank Approximation.} Before diving into the proposed procedure, it is useful to briefly review the basics of low rank assumption. Consider any matrix $\bA \in \mathbb{R}^{c_1\times c_2}$ having rank $rank(\bA)=r_A$ and nuclear norm $\|\bA\|_*$, the minimization of $r_A$ subject to convex constraints is known to be intractable. A typical remedy for this is to firstly assume $\bA$ is low rank, i.e., $r_A\ll \min({c_1,c_2})$, and then to relax the original rank minimization by the nuclear norm minimization using  $\|\bA\|_*$, since $\|\bA\|_*$ is provably the tightest convex approximation of $r_A$ under the low rank assumption \cite{candes2011robust}. While $r_A$ counts the actual number of nonzero singular values of $\bA$, nuclear norm $\|\bA\|_*$ computes the exact sum of the singular values of the matrix. A close analogy is how $\ell_1$ has replaced $\ell_0$ norm in the context of sparse coding, where $\ell_0$ counts the number of nonzero elements in a vector, and $\ell_1$ is the sum of magnitude of nonzero elements in a vector.

\subsection{Contrastive Learning with Low Rank Prior}

\begin{figure}[t]
 \centering
 \subfigure[]{\label{fig:cl}
  \includegraphics[width=0.055\textwidth]{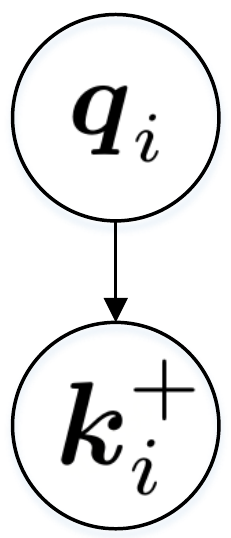}}\hspace{0.2cm} \hspace{0.5cm}
 \subfigure[]{\label{fig:gcl}
  \includegraphics[width=0.155\textwidth]{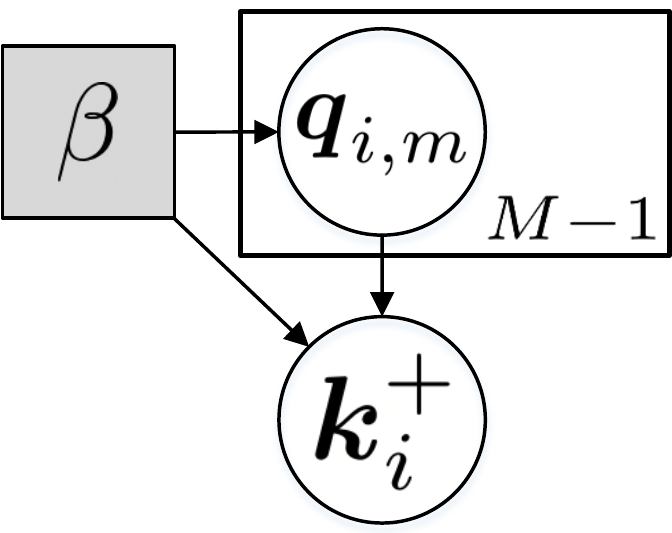}} \hspace{-0.2cm}
 \caption{\small Probabilistic graphical model on query-key pairs respectively for: (a) positive pair dependency based on conventional contrastive learning, (b) positive pair dependency based on LORAC.}
 \label{fig:PGM}
\end{figure}

We begin with our motivation and hypothesis. Eq. (\ref{eq:moco}) manifests that the contrastive loss is globally minimized iff. $\bq^T_i \bk_i^+$ is maximized while $\bq^T_i\bk_{i,j}^-$ is minimized. We attempt to better quantify this conventional learning assumption (Fig. \ref{fig:cl}) by incorporation of the low rank hypothesis (Fig. \ref{fig:gcl}), which simultaneously favors correlation among all the positive keys of the same instance. Our proposal involves a LOw RAnk promoting prior for Contrastive learning (LORAC), a natural evolution of conventional contrastive learning with additional probabilistic prior.

\textbf{The Probabilistic Graphical Model.} We now formally introduce LORAC, a novel probabilistic graphical model (PGM) governed by a set of low rank driven parameters. For each instance $i$, we first generate $M-1$ augmented views $\bx^q_{i,m}, ~(m \in [1, M-1])$ and the corresponding $M-1$ query features $\bq_{i,m}=f(\bx_{i,m}^{q}, \btheta_q),~~(m \in [1, M-1])$, with an additional augmentation $\bx^k_{i}$ and the associated positive key feature $\bk^+_{i}=g(\bx_i^{k}, \btheta_k)$. We then concatenate these $M-1$ query feature $\bq_{i,m}$ and the positive key features $\bk^+_{i}$ as rows of the matrix $\bQ\in {\mathbb R} ^{M \times d } = [\bq_{i,1}, ...\bq_{i,m},...,\bq_{i,M-1},\bk^+_i]^T$. Our hypothesis is that the rank of $\bQ$ is expected to be relatively small in comparison to the dimension of $\bQ$, i.e., $d$ and $M$. Such assumption poses regulatory effect that requires $\bq_{i,m}$ and $\bk^+_{i}$ of the $i$ instance to be generated from the same subspace.

Under such construction of $\bQ$ matrix, our objective is to maximize the joint probability of the PGM in Fig. \ref{fig:gcl} with regard to $\btheta_q$ and $\btheta_k$. Interestingly, this can be shown equivalent to maximizing the product of conditional distribution in correctly classifying the positive key $\bk^+_i$, given various queries $\bq_{i,m}$ and $\bQ$. For each instance $i$, we therefore equivalently maximize the product of the conditional distributions:
\begin{align}
\small p^* =\prod_{m=1}^{M-1} p(v[ \bk^+_{i}]= 1 |X,\bq_{i,m},\bQ),
\label{eq:gclPGM1}
\end{align}
where each $p(v[\bk^+_{i}]=1 |X,\bq_{i,m},\bQ)$ is defined as:
\begin{align}
&~~~\footnotesize{p( v[ \bk^+_{i}]= 1 |X,\bq_{i,m},\bQ)=} \nonumber \\
&\scriptsize{\frac{p(\bk^+_{i}|\bq_{i,m}, \bQ)\prod_{j=1}^K p(\bk^-_{i,j})} {p(\bk^+_{i}|\bq_{i,m}, \bQ)\prod_{j=1}^Kp(\bk^-_{i,j})+  \sum_{j=1}^{K} p(\bk^-_{i,j}|\bq_{i,m},\bQ)p(\bk^+_{i})\prod_{\ell \neq j}^K p(\bk^-_{i,\ell}) }}.
\label{eq:gclPGM2}
\end{align}
The equation above is daunting. However, if one takes a closer inspection at Eq. (\ref{eq:gclPGM2}), the only modification in comparison to Eq. (\ref{eq:cpc}) so far is essentially the replacement of the conditional probability $\bp(\bk^+_{i}|\bq_i)$ and $\bp(\bk^-_{i,j}|\bq_i)$ in Eq. (\ref{eq:cpc}) by the extra parameterization using conditionals $p(\bk^+_{i} |~\bq_{i,m},\bQ)$ and $p(\bk^-_{i,j} |~\bq_{i,m},\bQ)$. Note that we assume the negative keys $\bk^-_{i,j}, ~j \in [1,K]$ are statistically independent on each other as well as on positive key, for they are intrinsically sampled from different distributions. Therefore we have $p(\bk^-_{i,j}|~\bq_{i,m}, \bQ)=p(\bk^-_{i,j}|~\bq_{i,m})$, and Eq. (\ref{eq:gclPGM2}) correspondingly reduces to:
\begin{align}
&\footnotesize p(v[\bk^+_{i}]=1 |X,\bq_{i,m},\bQ)=\nonumber \\
&~~~\scriptsize \frac{{p(\bk^+_{i}|\bq_{i,m},\bQ)}/{p(\bk^+_{i})}}{{p(\bk^+_{i}|\bq_{i,m},\bQ)}/{p(\bk^+_{i})}+\sum_{j=1}^{K} {p(\bk^-_{i,j}|\bq_{i,m})}/{p(\bk^-_{i,j})}}.
\end{align}
Similar to the incentive of Eq. (\ref{eq:density:cpc}), our line of reasoning proceeds by looking for some functions that hopefully characterize the new density ratio. Specifically, we model:
\begin{align}
\small \frac{p(\bk^+_{i}|\bq_{i,m},\bQ)}{p(\bk^+_{i})} &\small \propto f_1(\bk^+_{i}, \bq_{i,m},\bQ) \nonumber \\
&\small = \exp({\bq_{i,m}^T \bk_i^{+}}/\tau) \cdot   h(\bQ), \label{eq:fkqQ2}\\
\small
\frac{p(\bk^-_{i,j}|\bq_{i,m})}{p(\bk^-_{i,j})}  & \small \propto   f_2(\bk^-_{i,j}, \bq_{i,m})= \exp ({{\bq_{i,m}^T \bk_{i,j}^{-}}}/{\tau}).\label{eq:fkqQ3}
\end{align}
There exist many options to perform the function $h(\bQ)$. We herein tentatively propose to use the hypothesis defining function $h(\bQ)$ as a scaled Laplace distribution:
\begin{equation}
\label{eq:h}
\small
h(\bQ)=\exp({{- \|\bQ\|_*/  (M\cdot\beta\cdot \tau)}}),
\end{equation}
where $\beta> 0$ is a predefined hyperparameter that controls the variance of the prior, $M$ normalizes the dimensional effect in $\|\bQ\|_*$, and $\tau$ is simply the conventional temperature \cite{hinton2015distilling} that makes $h(\bQ)$ scalable in the softmax.

There are certainly some inductive bias introduced via Eq. (\ref{eq:h}). But the principle behind using this heuristic is primly because, provided that $\frac{p(\bk^+_{i}|\bq_{i,m},\bQ)}{p(\bk^+_{i,j})}$ still retains the mutual information $I(\bk^+_{i}|\bq_{i,m})$ as InfoNCE \cite{oord2018cpc} used in Eq. (\ref{eq:density:cpc}), the function $f_1(\bk^+_{i}, \bq_{i,m},\bQ)$ in Eq. (\ref{eq:fkqQ2}) is just a constrained version of mutual information conditioned on hypothesis $h(\bQ)$, while the scaled Laplace prior penalizes $ \|\bQ\|_*$ at an exponential scale. As $ \|\bQ\|_*$ reduces, $h(\bQ)$ naturally encourages strong similarity among all views constructing $\bQ$. We argue that, the option of scaled Laplace distribution is nothing unique. In principle, all sparse priors of interest can be potentially considered in this manner, including the Jeffreys, Student’s t, and Gaussian priors. In practice (See Section \ref{sec:exp}), we find Laplace distribution empirically the best when LORAC serves as a pre-training approach, although alternative prior options are also adequate for obtaining good results.

\begin{figure}[t]
	\centering
	\includegraphics[height=0.2\textwidth]{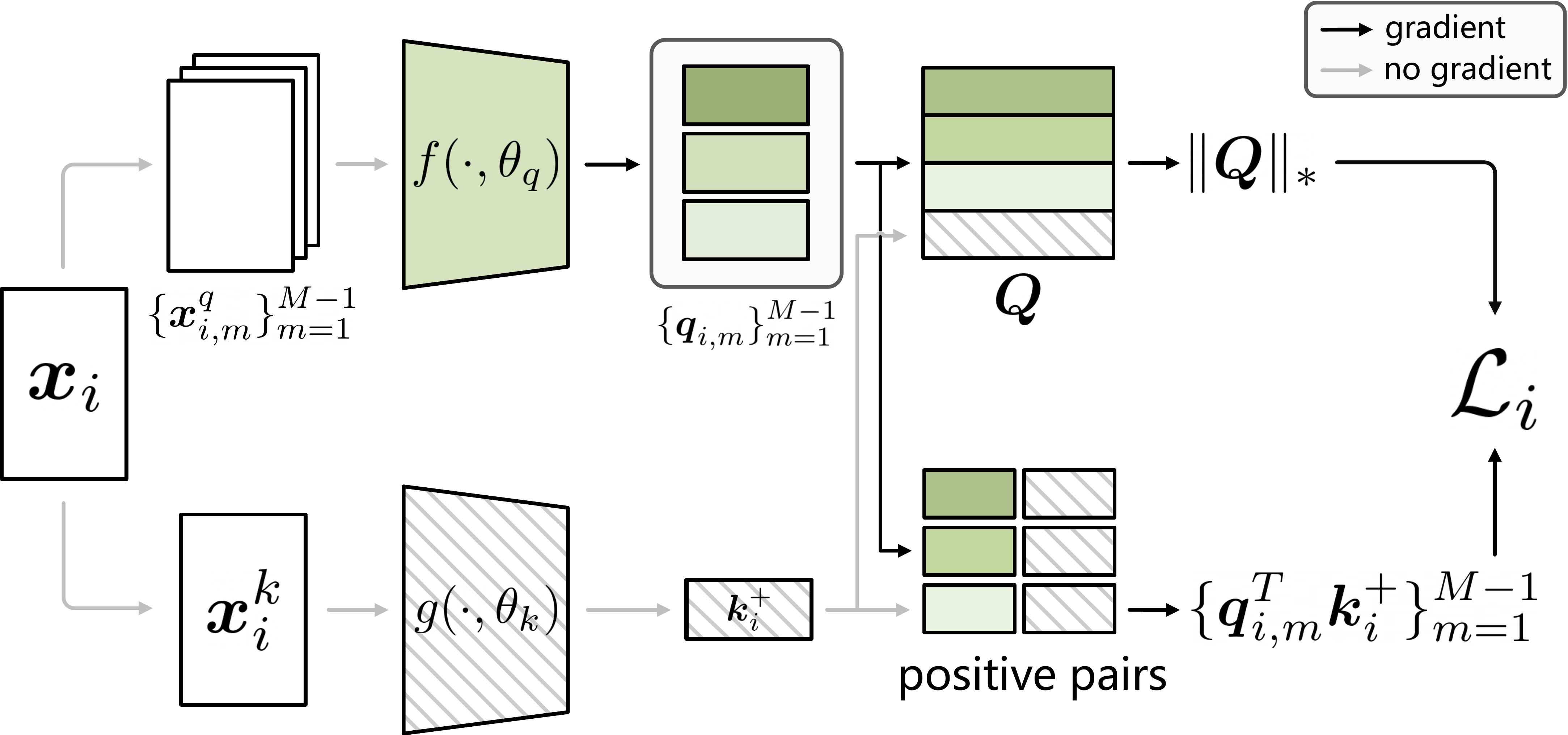}
     \vspace{-0.1cm}
	\caption{\small Conceptual framework of LORAC training procedure.}
	\label{fig:framework}
\end{figure}

\textbf{Overall Loss Function.} To summarize, the objective is to maximize the probability $p^*$ specified in Eq. (\ref{eq:gclPGM1}), with the $f_1$ and $f_2$ defined in Eq. (\ref{eq:fkqQ2}) and Eq. (\ref{eq:fkqQ3}). By applying $-\log$ to Eq. (\ref{eq:gclPGM1}), and normalizing by $M-1$, the loss function of the $i^{\text{th}}$ instance reduces to the minimization of:
\begin{equation}
\scriptsize
	\mathcal{L}_{i}=- \frac{1}{{\tiny M-1}}\sum_{m=1}^{M-1}\log \frac{\exp({{ \bq_{i,m}^T \bk_i^{+}}/\tau})    \cdot  h(\bQ) }  {\exp({{ \bq_{i,m}^T \bk_{i}^{+}}/ \tau})    \cdot h(\bQ)  +\sum_{j=1}^{K} {\exp({\bq_{i,m}^T \bk_{i,j}^{-}/\tau})}},
	\label{eq:Liloss}
\end{equation}
where $h(\bQ)$ is defined according to Eq. (\ref{eq:h}). The overall batch-wise loss during each training iteration boils down to:
\begin{equation}
\label{eq:finalloss}
\small
\mathcal{L}= \frac{1}{N}\sum_{i=1}^N\mathcal{L}_{i},
\end{equation}
where $N$ is the batch size, and $i$ indexes over instances.

\textbf{Implementation.} Recall that for MoCo, the query representation $\bq_i$ is obtained out of the encoder network $f(\cdot, \btheta_q)$, and $\bk_{i}^+$ comes from the key encoder network $g(\cdot, \btheta_k)$. The parameters $\btheta_q$ are updated via backpropagation, whereas the parameters $\btheta_k$ are only momentum updated versions of the $\btheta_q$ parameters without backpropagation.

We make mild modifications to these MoCo implementations in order to facilitate the optimization of loss Eq. (\ref{eq:Liloss}) and Eq. (\ref{eq:finalloss}). LORAC training procedure is illustrated in Fig. \ref{fig:framework}. We now have multiple $\bq_{i,m}$ to backpropagate through $f(\cdot, \btheta_q)$, while the gradients through $\bk_i^+$ and $\bk^-_{i,j}$ are frozen, and $g(\cdot, \btheta_k)$ is only momentum updated. This ``multiple query'' set-up is critical for LORAC to effectively work, as the additional $\|\bQ\|_*$ constraint relies on the active backpropagation through multiple features. If we had alternatively constructed $\bQ$ using multiple $\bk_i^+$ vectors instead of $\bq_{i,m}$, the effect of the $h(\bQ)$ prior would be infinitesimal, owing to the frozen backpropagation via $\bk_i^+$ and the slowly evolving $g(\cdot, \theta_k)$ that lags behind. In regard of the negative keys, we adopt MoCo's memory bank construction on $\bk^{-}_{i,j}$, so that LORAC is accessible to as many negative keys as possible. One intriguing phenomenon behind Eq. (\ref{eq:Liloss}) is that, under the scenario of $\beta \rightarrow \infty$, we obtain a plain implementation of multi-query contrastive learning without any low rank prior (equivalent to removal of $\|\bQ\|_*$ in Fig. \ref{fig:framework}), which we denote as MoCo-M. Here ``-M'' is reminiscent of ``multiple'', and ``MoCo-M'' serves as a constructed MoCo baseline with multiple positive views. In order to highlight the exclusive benefit offered by $h(\bQ)$, we compare the behaviors between MoCo-M and LORAC in Section \ref{sec:exp}.

For the entire training procedure, please see Algorithm \ref{alg:LORAC}. LORAC can be conveniently optimized via end-to-end backpropagation training. There are in total 3 hyperparameters required: number of views $M$, the temperature $\tau$ defined as in MoCo, and prior variance $\beta$ elusively defined for LORAC. In practice, we find $\beta$ lies in a wide effective range as shown in Section \ref{sec:exp}. We well abate the effect of these hyperparameters in the experiment section.

\section{Analysis}\label{sec:anl}
It is necessary to describe some analytical properties that justifies LORAC's validity for its application to contrastive learning. To ease the discussion, we firstly rewrite the loss into:
\begin{equation}
\scriptsize
\mathcal{L}=\frac{1}{M-1}\sum_i \log \left[ 1+\sum_j^K \exp\left(\bq_{i,m}\bk_{i,j}^-/\tau-\bq_{i,m}\bk_{i}^+/\tau+h(\bQ) \right)  \right].
\label{eq:discussion}
\end{equation}
It manifests now that, without $h(\bQ)$, original contrastive loss aims to minimize the contrastive term $\bq_{i,m}\bk_{i,j}^-/\tau-\bq_{i,m}\bk_{i}^+/\tau$ alone. In other words, contrastive loss aims to independently maximize the difference between $\bk_{i,j}^-$ and $\bk_{i,j}^-$ along each of the direction of $\bq_{i,m}$. In the meanwhile, the net effect of LORAC imposes an additional prior $h(\bQ)$ that further requires all $\bq_{i,m}$ should have an interplay of each other and jointly lie in a low dimensional subspace, forcing the dependency among all views. Also, independent from the probabilistic view of the proposed model though, it is immediate that any increase in $\|\bQ\|_*$ necessarily and monotonically increases $\mathcal{L}_i$ in Eq. (\ref{eq:Liloss}). Therefore, the $\bQ$ prior in Eq. (\ref{eq:h}) effectively penalizes any increase in $\|\bQ\|_*$. According to \cite{Boyd2001rank}, penalizing nuclear norm effectively decreases the convex envelope of $rank(\bQ)$\footnote{Notably, the equivalence between the $rank$ and nuclear norm is sensitive to the data correlation and matrix dimension.}. In the meanwhile, negative pair terms effectively prevent the value of $\|\bQ\|_*$ from going to zero, to avoid collapsed solution that all representations of all samples are mapped to the same point. Particularly, in regard of the global minimum of Eq. (\ref{eq:Liloss}), we have the following observation (see supplementary material for proof):

\begin{theorem}
Given normalized features $\bq^*_{i,m}$ and $\bk^*_{i}$, if $h(\bQ)= \exp(-(\|\bQ\|_*-c_o)^2/(M\cdot\beta\cdot\tau)) $, where $c_o=\sqrt{M}$, there exists a $\beta > 0$ sufficiently large such that a local optimum $\bq^*_{i,1},...,\bq^*_{i,m},...,\bq^*_{i,M-1},{\bk^*}^+_i$ of Eq. (\ref{eq:finalloss})  lie on the same low-dimensional subspace for each $i$. This local optimum returns the same loss value as local minima of MoCo-M (Eq. (\ref{eq:finalloss}) with $h(\bQ)=0$), and conventional contrastive learning (Eq. (\ref{eq:finalloss}) with $h(\bQ)=0$ and $M=2$).
\label{theo:1}
\end{theorem}

The first implication of Theorem \ref{theo:1} is that with particular form of prior $h(\bQ)$, LORAC is guaranteed to be anchored at some local mimima that is as good as conventional contrastive learning (with $h(\bQ)=0$) if there are multiple query views present. In this regard, at least LORAC under this particular set-up would not do worse than conventional contrastive learning. In this case, the advantage of LORAC really is smoothing away additional bad local minimas than MoCo, via the reduced degree of freedom of the optimization problem for network parameters. To see this, note that for conventional contrastive learning (Eq. (\ref{eq:discussion}) with $h(\bQ)=0$), every independent $\bq_{i,m}$ having $\sum_j (\bk^+_{i}-\bk^-_{i,j})\exp(\bq_{i,m}^T \bk^-_{i,j})=0$ necessarily create a local minimum of the loss (zero gradient through $\bq_{i,m}$) with regard to the network parameter, even if $\bq_{i,m}$ for distinct $m$ might drastically deviate from each other. In contrast, LORAC loss can well escape these local minima via penalizing $h(\bQ)$ in order to further align different $\bq_{i,m}$ for distinct $m$, without increasing the contrastive loss between query-key pairs than conventional contrastive learning (Eq. (\ref{eq:discussion}) with $h(\bQ)=0$ and $M=2$) according to Theorem \ref{theo:1}. In other words, LORAC can reach a local minimum of loss Eq. (\ref{eq:Liloss}) by further penalizing $h(\bQ)$, {\emph{without increasing the optimal contrastive loss value between sample pairs}}.

\begin{algorithm}[t]

      \caption{LORAC: PyTorch-like Pseudocode}

      \scriptsize

      \begin{algorithmic}

      \ttfamily

           \State{\color{PineGreen} \# f: encoder f}

           \State{\color{PineGreen} \# g: encoder g}

           \State{\color{PineGreen} \# queue: dictionary queue (C,K)}

           \State{\color{PineGreen} \# N: batch size}

           \State{\color{PineGreen} \# M: number of views}

           \State{\color{PineGreen} \# tau: temperature}

           \State{\color{PineGreen} \# beta: prior variance of LORAC}

           \State{}

           \State{{\fontseries{b}\selectfont \color{Magenta}for} x {\fontseries{b}\selectfont \color{Magenta}in} loader{\color{Magenta}:} {\color{PineGreen}\# load a mini-batch x (N samples)}}

           \State{\ \ \ \ \color{PineGreen} \# M random augmentations}

           \State{\ \ \ \  multi\_x0 {\color{Magenta}=} {\color{Magenta}[}aug(x) {\fontseries{b}\selectfont \color{Magenta}for} \_ {\fontseries{b}\selectfont \color{Magenta}in} {\fontseries{b}\selectfont \color{Magenta}range}(M-1){\color{Magenta}]}}

           \State{\ \ \ \   x1 {\color{Magenta}=} aug(x)}

           \State{\ \ \ \  \color{PineGreen} \# extract multi-query and key}

           \State{\ \ \ \   multi\_q {\color{Magenta}=} f(multi\_x0)        {\color{PineGreen}\# (M-1,N,C)}}

           \State{\ \ \ \  k {\color{Magenta}=} g(x1).detach()        {\color{PineGreen}\# (1,N,C)}}

           \State{\ \ \ \  \color{PineGreen} \# construct Q matrix}

           \State{\ \ \ \   Q {\color{Magenta}=} cat({\color{Magenta}[}multi\_q, k{\color{Magenta}]}, dim=0)        {\color{PineGreen}\# (M,N,C)}}

           \State{}

           \State{\ \ \ \   loss {\color{Magenta}=} L(multi\_q, k, queue, Q)  {\color{PineGreen}\# Eq.(13)} }

           \State{\ \ \ \  loss.backward()}

           \State{}

           \State{\ \ \ \ update(f, g)}

           \State{\ \ \ \ queue.push(k)}

           \State{}

           \State{\color{PineGreen} \# contrastive loss with low rank prior}

           \State{{\fontseries{b}\selectfont \color{Magenta}def} L(multi\_q, k, queue, Q){\color{Magenta}:}}

           \State{\ \ \ \ loss {\color{Magenta}=} 0.}

           \State{\ \ \ \ {\fontseries{b}\selectfont \color{Magenta}for} q {\fontseries{b}\selectfont \color{Magenta}in} multi\_q{\color{Magenta}:} }

           \State{\ \ \ \ \ \ \ \ \color{PineGreen} \# h(Q): here nuclear norm}

           \State{\ \ \ \ \ \ \ \  rank {\color{Magenta}=} norm(Q, p{\color{Magenta}='nuc'}, dim{\color{Magenta}=}(0, 2))}.view(N,1)

           \State{\ \ \ \ \ \ \ \  rank {\color{Magenta}/=} M {\color{Magenta}*} beta}

           \State{}

           \State{\ \ \ \  \ \ \ \ \color{PineGreen} \# positive logits: (N,1)}

           \State{\ \ \ \ \ \ \ \  l\_pos {\color{Magenta}=} bmm(q.view(N,1,C), k.view(N,C,1))}

           \State{\ \ \ \  \ \ \ \ \color{PineGreen} \# negative logits: (N,K)}

           \State{\ \ \ \ \ \ \ \  l\_neg {\color{Magenta}=} mm(q.view(N,C), queue.view(C,K))}

           \State{\ \ \ \  \ \ \ \ \color{PineGreen} \# logits with low rank prior: (N,1+K)}

           \State{\ \ \ \ \ \ \ \   logits {\color{Magenta}=} cat({\color{Magenta}[}l\_pos {\color{Magenta}-} rank, l\_neg{\color{Magenta}]}, dim=0)}

           \State{}

           \State{\ \ \ \  \ \ \ \ \color{PineGreen} \# contrastive loss with low rank prior: Eq.(12)}

           \State{\ \ \ \ \ \ \ \  labels {\color{Magenta}=} {\fontseries{b}\selectfont \color{Magenta}zeros}(N)}

           \State{\ \ \ \ \ \ \ \  loss {\color{Magenta}+=} CrossEntropyLoss(logits {\color{Magenta}/} tau, labels)}

           \State{\ \ \ \ {\fontseries{b}\selectfont \color{Magenta}return} loss {\color{Magenta}/} (M-1)}

      \end{algorithmic}

      \label{alg:LORAC}

\end{algorithm}

Another useful practical implication behind Theorem \ref{theo:1} is that the density $\beta$ should be chosen appropriately. In fact,  $\beta$ reflects the variance of the hypothesized singular value distribution. As $\beta$ becomes sufficiently small, the optimization w.r.t. $\|\bQ\|_*$ dominates the learning procedure and makes the impact of the contrastive term ${\exp({\bq_{i,m}( \bk_{i,j}^--\bk_{i}^+ )/\tau}})$ in Eq. (\ref{eq:discussion}) infinitesimal. If this happens, the loss has no motivation to learn any contrastive features, and all of the instances in the training set would be mapped to an identical point in order to reduce $\|\bQ\|_*$. On the other hand, if $\beta$ becomes too large, the effect of $\|\bQ\|_*$ constraint vanishes, e.g.,  as $\beta$ approaches infinity, Eq. (\ref{eq:Liloss}) collapses to a summation over conventional InfoNCE loss Eq. (\ref{eq:moco}) without the coupling effect through $\bQ$. Under these considerations, we find it helpful to gradually anneal the value of $\beta$ as the training proceeds. This treatment directly echos our hypothesis that $\beta$ serves as the statistical variance of the low rank prior. At earlier training, a bigger $\beta$ relaxes the variance constraint on $\|\bQ\|_*$ and assigns a relatively flat prior on the choice of $\bQ$, owing to lack of certainty in the loss landscape. In this way, the loss poses less preferences on $\bQ$, so that the network focuses on learning contrastive features. As the training produces finer features, a decayed $\beta$ gradually assigns more probability density in the search area of low $\|\bQ\|_*$, so that intra-instance invariance becomes more favorable. Note that an extra bonus in exchange of this $\beta$ tuning is that, LORAC completely waives the critical Gaussian distributional assumptions on $\bk_i$ required by JCL \cite{cai2020joint} like algorithms. Therefore LORAC generalizes to a broader class of feature generative assumptions independent from their generative process distributional forms, i.e., LORAC is not tied to any local sufficient statistics of samples (crystal clear from the loss itself though), as long as the features respect the low rank regularization under any preferred $h(\bQ)$ in mind (we disambiguate in supplementary file).

\section{Extensions and Generalization} \label{sec:extension}
We have derived the prototype LORAC from a pure probabilistic point of view. In practice, in order to best respect the low rank assumption on views, the whole LORAC procedure with regard of $\bQ$ construction can be also adapted with flexibility given different view generation policies. Take for instance, when constructing $\bQ$ matrix, the effect of outliers (out of distribution samples) should be taken into account and handled carefully, due to similar concerns raised in RPCA \cite{candes2011robust}. Especially when there are prior knowledge that some type of views are generated with high variance that likely deviate from the principle component from other views (e.g., extremely small crops likely not capturing shared object, low resolution views w.r.t. other views, etc.), it is potentially helpful to remove these views from $\bQ$, while retaining them in the remaining part of the loss Eq. (\ref{eq:Liloss}). This effectively reduces the risk of out of distribution samples contribution to the principle component of the views computed via $\bQ$.

LORAC can also be extended to a closely relevant but alternative ``{\emph{batchwise low rank assumption}}''. Particularly, we assume that the ``variations'' from all views of all instances lie in the same subspace. This assumption is based on the fact that all self-supervised views are generated from the same augmentation pipeline and policy with varying hyperparameters. Therefore, the variations of all views around each instance mean are considered as linear combination of some basis representing augmentation actions such as rotation, crop, color jittering and etc factors, and those basis span the whole augmentation subspace. In this scenario, LORAC loss can be generalized and reduced into:
\begin{equation}
\scriptsize
	\bar{\mathcal{L}}_{i}=- \frac{1}{{\tiny M-1}}\sum_{m=1}^{M-1}\log \frac{\exp({{ \bq_{i,m}^T \bk_i^{+}}/\tau})    \cdot  h(\bP) }  {\exp({{ \bq_{i,m}^T \bk_{i}^{+}}/ \tau})    \cdot h(\bP)  +\sum_{j=1}^{K} {\exp({\bq_{i,m}^T \bk_{i,j}^{-}/\tau})}},
	\label{eq:barLiloss}
\end{equation}
where matrix $\bP \in \mathbb R^{(M-1)N\times d}$ is now shared across all instances $i\in [1,N]$ in the batch and constructed as $\bP=[\ba_{i,m}, ...\ba_{i,M-1}, ...\ba_{N,M-1}]^T$. Here, $\bP$ is a concatenation of centered views $\ba_{i,m}$ from each instance: $\bmu_i=\frac{1}{M-1}\sum_m^{M-1} \bq_{i,m}$, and $\ba_{i,m}= \bq_{i,m}-\bmu_i$. and we need to also normalize the matrix dimension via: $h(\bP)=\exp({{- \|\bP\|_*/  N(M-1)\cdot \beta\cdot \tau)}})$.

An optimization with regard to $\bar{\mathcal{L}}_i$ in Eq. (\ref{eq:barLiloss}) can efficiently save the computational cost of original LORAC loss, by reducing the SVD computing load to only once per batch with complexity $\mathcal O(d^3)$. However, we notice a performance drop when implementing $\bar{\mathcal{L}}_i$ with $\bP$ than training LORAC loss in Eq. (\ref{eq:Liloss}) with $\bQ$, in exchange of faster computation. We will investigate the difference of these LORAC variations via apple to apple empirical evaluation in Section \ref{sec:exp}.

\section{Experiments}\label{sec:exp}
In this section, we evaluate the proposed hypothesis via various empirical evidence, and we compare the proposed LORAC with the state-of-the-art unsupervised learning approaches, including several contrastive learning paradigms. Recently, checking whether or not pre-trained features can generalize well becomes a typical evaluation for unsupervised/supervised feature learning \cite{Swav2020,he2019momentum,he2016deep,li2021contextual}. We thus prioritize to compare the algorithms under unsupervised setting. According to \cite{Swav2020,he2019momentum}, we consider the following evaluations when compare the quality of pre-trained features obtained from various algorithms: (a) Linear classification on ImageNet1K dataset \cite{deng2009imagenet}, directly trained on frozen pre-trained features; (b) Transfer to classification tasks on other domains and semi-supervised learning tasks; (c) Finetuning the networks initialized from the pre-trained network for downstream tasks, including object detection, instance segmentation and keypoint detection on MS COCO dataset \cite{lin2014microsoft}; (d) Ablation studies that support our assumptions on the proposed model.

\begin{table*}[t]
	\caption{Published/reimplemented accuracy of linear classification on features pre-trained on ImageNet1K. Symbol $^\#$ are results borrowed from \cite{kolesnikov2019revisiting}, showing better performances than original publications;  $^\S$ are the results of our reimplemented model based on code downloaded from MoCo-v2 website \href{https://github.com/facebookresearch/moco}{https://github.com/facebookresearch/moco}, and from SwAV website \href{https://github.com/facebookresearch/swav}{https://github.com/facebookresearch/swav}  }	
	\begin{center}
		\tablestyle{9pt}{1.2}
		\begin{adjustbox}{max width=1\textwidth}
		\begin{tabular}{l|llllll|ll}
			\shline
			Method               &  batchsize   &                train. epoch & multi-crop & num. forward      & arch.  & pars. (M) & Top-1            & Top-5 \\
			\hline
			Jigsaw \cite{noroozi2016unsupervised}    & 256 & 69 & /& 1  & ResNet-50(2x) & 94         & 44.6$^\# $             & 68.0$^\# $ \\
			Rotation \cite{gidaris2018unsupervised}  & 192  & 30 & / & 1  & RevNet(4x)    & 86         & 55.4$^\# $             & 77.9$^\# $ \\
			DeepCluster \cite{caron2018deep}     &   256 & 500   &/  &  1 & VGG           & 15         & 48.4                        & \slash          \\
			BigBiGAN \cite{donahue2019large}   &  2048  &  1569  &/ &  1   & RevNet(4x)    & 86         & 61.3                        & 81.9            \\
			\hline
			CPC-v1 \cite{oord2018cpc}            &  512 &  200  & / &  1  & ResNet-101    & 28         & 48.7                        & 73.6            \\
			CPC-v2 \cite{henaff2019data}        &  512  &  200  & / &  1  & ResNet-50     & 24         & 63.8                        & 85.3            \\
			CMC \cite{tian2019contrastive}      &  128  &  200  & 2 &  2  & ResNet-50     & 47         & 64.0                        & 85.5            \\
			SimCLR \cite{chen2020simple}      &  4096  &  1000   & 2$\times$224 &   2  & ResNet-50     & 24         & 69.3 &   89.0              \\
			JCL  \cite{cai2020joint}                    &   256 &  200         & 5$\times$224  &   5    &ResNet-50   & 24& 68.7 &89.0 \\
			MoCo-v1\cite{he2019momentum} &   256 &   200   & 2$\times$224 &     2      & ResNet-50     & 24         & 60.6  & 83.1$^\S$       \\
			MoCo-v2 \cite{chen2020improved}  & 256 &  200  &  2$\times$224 &    2   & ResNet-50     & 24         & 67.5$^\S$ & 88.0$^\S$       \\
			MoCo-v2+\cite{chen2020improved}  &  256&  800  & 2$\times$224 &   4    & ResNet-50     & 24         & 72.2$^\S$  &   /     \\
                BYOL\cite{BYOL2020}                                              &  4096  & 400  &  2$\times$224  & 4 & ResNet-50  &24 &{\bf 73.2} & /\\
                Barlow Twins \cite{barlowtwins2021}                                 &   2048  &1000 &   2$\times$224  & 2 & ResNet-50  &24 &{\bf 73.2} &{91.0} \\
			\hline			
			SwAV   \cite{Swav2020}                                         & 256       &   200      &    2$\times$224 + 6$\times$96    &10 & ResNet-50  &24 &{72.7}$^\S$ & {91.5}$^\S$ \\
			SwAV     \cite{Swav2020}                                       & 256       &   200      &    3$\times$224 + 5$\times$96    & 10 & ResNet-50  &24 &{72.3}$^\S$ &{91.1}$^\S$ \\
			MoCo-M                                       &256        &  200       &  3$\times$224 + 5$\times$96   & 10 & ResNet-50  &24 &{72.5} & 91.3 \\
 			LORAC                                        & 256       &   200      &   3$\times$224 + 5$\times$96     & 10 & ResNet-50  &24 &{\bf 73.2} & \bf 91.6 \\
			\shline
		\end{tabular}
		\end{adjustbox}
	\end{center}
	\label{tab:imagenet_main}
\end{table*}

\subsection{Pre-training on ImageNet1K}\label{sec:evllc1}
{\bf {Dataset and Optimization Setups:}} In all of the following sections, all the algorithms presented have been pre-trained on the ImageNet1K dataset \cite{deng2009imagenet}. For LORAC and MoCo-M pre-training, we use ResNet-50 \cite{he2016deep} as the backbone network. An additional two-layer MLP (hidden layer 2048-$d$ with ReLU) is added on top of the global pooling layer of ResNet-50, and the final dimension of embedding is $d=128$. The batch size is $N=256$, which is distributed on 4 GPUs. We employ the LARS \cite{you2017large} optimizer with base learning rate $lr=1.8$, final learning rate $0.018$ upon cosine decay rule. We train LORAC for 200 epochs on ImageNet1K dataset to pretrain the ResNet-50 network. Following the analysis in Section \ref{sec:anl}, we train LORAC in a step-wise schedule where $beta$ decays at certain epochs. Specifically, we use $\beta=\infty$ during the $1^{th}- 100^{th} $ epoch, and $\beta=4$ for $101^{th}- 200^{th} $ epochs.  For other hyperparameters in loss Eq. (\ref{eq:Liloss}), $\tau=0.2$. The $h(\bQ)$ function follows Eq. (\ref{eq:h}) through out this Section \ref{sec:evllc1}.

Particularly, as our ultimate goal is to leverage the dependency across multiple views, we borrow the data augmentation policy called ``multi-crop'' from SwAV to apply on MoCo-M and LORAC, for SwAV is a typical state-of-the-art multi-view training algorithm. Specifically, we choose to use the 3$\times$224 + 5$\times$96 multi-crop augmentation to apply on MoCo-M and LORAC. According to the ``multi-crop'' strategy in SwAV, during each training iteration, we generate 3 full views of $224\times224$ and 5 additional views of $96\times96$ (i.e., 8 views in total, and 10 forward per iteration as defined in SwAV) with scale factor ranging in $[0.14, 1.0]$ (for $224\times224$ views) and in $[0.05, 0.14]$ (for $96\times96$ views) respectively. We then apply color distortion augmentation, which includes color jitter (i.e., the strength of \{brightness, contrast, saturation, hue\} is \{$0.8, 0.8, 0.8, 0.2$\} with probability of $0.8$) and color dropping (i.e., conversion to grayscale with probability of $0.2$) to each view. Finally, Gaussian blur with a standard deviation in $[0.1, 2.0]$ is applied on top of the obtained augmentation views. We adopt the option of using relatively large views ($3\times 224$ views) to construct $\bQ$ to reduce risks of outliers, as per the discussion in Section \ref{sec:extension}. In the meanwhile, in order to compare against SwAV, we reproduce the SwAV pre-training performance by using both the published 2 $\times$224 + 6$\times$96 in \cite{Swav2020} and 3$\times$224 + 5$\times$96 strategy, the same as used for LORAC and MoCo-M. For MoCo-v2, the augmentations are generated in the same way as published in \cite{chen2020improved}, as the multi-crop augmentation is not applicable on MoCo-v2. Specifically, a 224$\times$224 patch is firstly randomly cropped from the resized images, then we apply color jittering, random grayscale, Gaussian blur, and random horizontal flip to each patch of MoCo-v2. We list the respective results in the following section.

\begin{table*}[t!]
 \caption{\small Linear classification on other tasks.} \vspace{-0.7cm}
 \begin{center}
  \tablestyle{6pt}{1.25}
  \begin{adjustbox}{max width=1\textwidth}
  \begin{tabular}{c | c c c c c c c c c c c c | c}
   \shline
  Model         &Aircraft     & Birdsnap     &       Caltech-101  &  Cars     &CIFAR10   &   CIFAR100    &DTD   &Flowers   & Food &      Pets   &  Place365 & SUN397 &Average  \\
  \hline
   Supervised    & 42.6 & \bf 57.2 & 90.9 & 62.8 & 90.0 & 73.4 & 68.8 & \bf 89.7 & 71.3 & \bf 92.4 & 49.1 & 60.3 & 70.7\\
   \hline
   MoCo-v1    &    34.2                     &  32.2  &    77.5   &   40.5          &  82.3    & 60.2     &  64.5   & 80.8 & 67.5  & 73.9 &    45.7  & 52.2 &59.3 \\
   MoCo-v2         &     39.3                    & 40.7   &  88.6     &   52.9          & 89.8     &  71.0    &  69.2   & 87.4 & 72.5 &82.9&    48.7       &59.2  &66.9\\
   \hline
   SwAV& 45.5 & 49.0 & 98.0 & 62.2 & 90.8 & 73.5 & 72.2 & 88.9 & 74.3 &  87.1 & 51.4 & 64.4 & 71.4 \\
   MoCo-M & 44.5 & 49.3 & 98.0 & 60.5 & 91.4 & 74.0 & 72.6 & 88.9 & 73.7 & 86.4 & 51.2 & 64.1 & 71.2 \\
   LORAC  & \bf 47.8 & 50.9 & \bf 98.1 &\bf  63.5 & \bf 91.8 &\bf  75.3 &\bf  72.7 &  89.5 & \bf 76.0 & 86.8 & \bf 51.9 & \bf 64.9 & \bf 72.4 \\
   \shline
  \end{tabular}
  \end{adjustbox}
\end{center}
\label{tab:12datasetsemi}
\vspace{-0.5cm}
\end{table*}

Correspondingly, to associate with the ``multi-crop'' augmentation, LORAC and MoCo-M both adopt the same augmentation ($3\times 224+5\times96$), pre-training and downstream optimization technique as SwAV, i.e., the LARS optimizer, cosine learning rate schedule and query key pairing policy among views. For LORAC and MoCo-M pre-training, start learning rate is 1.8. The optimizer momentum is 0.9, with a weight decay $10^{-6}$. We regard MoCo-M as a critical multi-view MoCo-v2 baseline that properly abates the effect of $h(\bQ)$ function.

{\bf Linear Classification on Frozen Features:}
The pre-trained feature quality directly affects the performance of downstream linear classification. We firstly evaluate the linear classification performance on frozen features obtained from SwAV, MoCo-M and LORAC according to the evaluation protocols defined in \cite{he2019momentum}. Given the pre-trained MoCo-v2 model, we use batch size $N=256$ and SGD optimizer with $lr=30$, momentum of $0.9$ to train the classifier, without weight decay. For model pre-trained using MoCo-M, LORAC, we use batch size $N=256$, SGD optimizer with starting $lr=1.2$ upon a cosine learning rate schedule.

Table \ref{tab:imagenet_main} compares classification accuracy with the state-of-the-art unsupervised methods. We list the published pre-training models with their performance on linear classification accuracy in Table \ref{tab:imagenet_main}. Particularly, we also consider the following factors: {\emph{model size, training epochs, number of views included per training iteration, and {\emph{number of forward during each training iteration}} (abbreviated as num. forward) }} so that approaches are compared in an apple to apple fashion. Readers are encouraged to pay attention to these factors, as those factors significantly affect the performance and properly take into account the fairness issue. We denote the MoCo-v2 approach having ``swap'' loss as MoCo-v2+ \cite{chen2020exploring}, as this variant of MoCo-v2 effectively doubles the number of forward. Especially, among all baselines, we are most interested in comparing with our constructed MoCo multi-view baseline, defined as MoCo-M (LORAC with $\beta=\infty$), as MoCo-M ablated the effect of $h(\bQ)$ against LORAC. To compare MoCo-M and LORAC in a fair and clean manner, MoCo-M pre-training has used exactly the same data augmentation policy, optimization and hyperparameters as for LORAC and SwAV. Especially, the readers can refer to the difference on number of forward and epochs to compare among approaches. Take for instance, LORAC has 10 forwards per training iteration, therefore should be compared with MoCo-v2+'s ($200\times 10/4=500$) epochs, although we find that LORAC (200 epochs) and MoCo-M (200 epochs) are even both providing a much better performance than MoCo-v2+ (800 epochs). Similarly, LORAC (200 epochs) with batchsize 256 even shows comparable performance to BYOL's (400 epochs) that requires batchsize 4096 and to Barlow Twins (1000 epochs) which requires batchsize 2048. Please note, both BYOL and Barlow Twins have claimed their sensitivity on batchsize (performance drops when batchsize decreases), whereas LORAC has achieved comparable performance (even higher top-5 score than Barlow Twins) without having to resort to huge batchsize.   This again shows that LORAC's exclusive optimization advantage over conventional approaches.

As Table \ref{tab:imagenet_main} shows, LORAC performs strong superiority over other comparable approaches. Firstly, the clear advantage of LORAC over its MoCo Multi-view baseline (i.e., MoCo-M) empirically justifies the effectiveness of proposal Eq.(\ref{eq:Liloss}) and corroborates our hypothesis throughout this paper. We believe the construction of MoCo-M baseline has properly abated the effect of multi-views, optimization, and augmentation policy issues against the $h(\bQ)$ term of LORAC. Still, with the introduced $h(\bQ)$ function, LORAC successfully smooths away bad local minimas that the independent pairwise training would produce, and correspondingly provides better instance level invariance and inter-view concentration than the conventional MoCo Multi-view baseline MoCo-M. Notably, LORAC also surpasses SwAV by $0.5\%$, the state-of-the-art multi-view training self-supervised learning method, well verifying the validity of our subspace assumption and superiority over the cluster assumption in finding good pre-training features. Even when comparing with MoCo-v2+ baseline (2 views, 4 forward) trained for 800 epochs, LORAC 200 epochs still retains clear performance advantage, again, strongly corroborating our hypothesis that joint training (MoCo-M and LORAC) helps. This shows that LORAC converges to a different local minimum and reaches higher performance in fewer epochs than MoCo's comparable epochs (10 forward/4 forward $\times$ 200epochs = 500 epochs).

{\bf Transfer to other Linear Classification Tasks:} Does LORAC simply overfit to Imagenet1K and lead to better performance on downstream Imagenet1K classification? To address this issue, we evaluate LORAC's generalization across various downstream tasks in different data domains as in \cite{chen2020simple,BYOL2020} (but we do not compare with their published results trained for 1000 epochs on ResNet50$(4\times)$). We follow the pre-train protocol as in Section \ref{sec:evllc1}, freeze the backbone network and only train classifier parameters on the following datasets:   Aircraft \cite{aircraft}, Birdsnap \cite{birdsnap},  Caltech-101 \cite{caltech101}, Cars \cite{CARSDATAset},     CIFAR10 \cite{krizhevsky2009cifar10and100},   CIFAR100 \cite{krizhevsky2009cifar10and100}, DTD \cite{DTDdataset},  Flowers \cite{flowersdata}, Food \cite{food101dataset}, Pets \cite{catsdataset}, Place365 \cite{places2014}  and SUN397 \cite{SUN397}. We use the same training hyperparameter and optimizers for all methods in Table \ref{tab:12datasetsemi}.

For fair comparisons, the MoCo-M, LORAC and SwAV approaches presented in Table~\ref{tab:12datasetsemi} are pre-trained for 200 epochs. We download the officially published MoCo-v2 model pretrained on ImageNet1K for 200 epochs as a baseline model. We compare with MoCo-M, LORAC and SwAV pretrained for 200 epochs. Following \cite{chen2020simple, BYOL2020}, we adopt {\bf mean per class} accuracy as evaluation protocol for Aircraft, Caltech-101, Flowers, Pets. In the meanwhile, we use {\bf top-1} accuracy as evaluation metric for the remaining datasets: Birdsnap,  Cars, CIFAR10, CIFAR100, DTD, Food, Places365 and SUN397. The ``supervised'' baseline is a fully supervised model pre-trained on ImageNet1K. The model is downloaded from  \href{https://download.pytorch.org/models/resnet50-19c8e357.pth}{https://download.pytorch.org/models/resnet50-19c8e357.pth}.

The image pre-processing during linear classification training and test time remains the same as for the linear evaluation on ImageNet1K. For all of experiments on transfer learning tasks, we train the linear classifiers on top of the frozen features out of the pre-trained network with an SGD optimizer. We use batch size $N=256$, momentum of $0.9$ and weight decay of $0$ for optimization. We search the optimal base learning rate for all algorithms in a range of 7 logarithomically-spaced values between $10^{-1}$ and  $10^{2}$, and we anneal the optimal learning rate for each algorithm by cosine decay rule. Due to the varying sizes among these 12 datasets, we adopt suitable training epochs / iterations for each dataset. Particularly, we train the linear classifiers for 100 epochs on datasets  Place365, and train 30,000 iterations for Birdsnap, Cars, CIFAR10, CIFAR100, Food and train 5,000 iterations for Aircraft , Caltech-101, DTD, Flowers, Pets. In addition, owning to the broken images found in Birdsnap, we had to firstly sample 5 random validation images from each category, and sample 5 random testing images per category. We then use the remaining 35,447 number of valid images for training.

As Table \ref{tab:12datasetsemi} shows, LORAC exhibits obviously better generalization than its MoCo-M baseline, when the pre-trained features are directly transferred to other downstream datasets. The results basically confirm that LORAC indeed benefits from a better instance level invariance via $h(\bQ)$, which effectively reduces the degree of freedom of the network optimization, as the advantage is well transferable to other datasets. Particularly, Place365 dataset has 365 classes and in total 1.8 million training images with more than 300,000 test images, which even includes much more training images than the pretraining ImageNet1K dataset. A better accuracy on this challenging dataset shows that LORAC indeed has changed the optimization landscape of the pre-training than its baselines, which changed the feature distribution for the downstream task. Even if the downstream task has enormous labeled data, the benefit offered by LORAC is not erased. Notably, all of the SwAV, MoCo-M, and LORAC have outperformed the supervised pre-training baseline, while LORAC has surpassed the supervised baseline.

\begin{table}[t!]
\caption{Semi-supervised learning evaluation as defined in \cite{Swav2020}. Accuracy in $\%$.}\vspace{-0.5cm}
    \centering
    \tablestyle{14pt}{1.2}
    \begin{adjustbox}{max width=0.495\textwidth}
    \begin{tabular}[t]{c|cc|cc}
        \shline
         \multirow{2}*{Model } & \multicolumn{2}{c}{\em1\% labels } & \multicolumn{2}{c}{\em 10\% labels }\\
         \cline{2-5}
          ~      &Top-1   &Top-5 &Top-1&Top-5\\ \hline
          MoCo-v1 & 29.8  &55.9 &49.2&75.2 \\
          MoCo-v2 & 44.0  &70.9 &58.8&82.8 \\
          \hline
        SwAV  & 49.6 & 76.1 & 67.7 & 88.7 \\
        MoCo-M & 48.4 &75.4 & 66.9 & 88.5 \\
        LORAC & \bf 50.0 & \bf 76.3 &\bf  68.0 & \bf 88.9 \\
        \shline
    \end{tabular}
    \end{adjustbox}
    \label{tab:semi}
\end{table}

\begin{table*}[t!]
\caption{Performance on downstream tasks: object detection \cite{ren2015faster} (left), instance segmentation  \cite{he2017mask} (middle) and keypoint detection \cite{he2017mask} (right). Accuracy in $\%$. All models pre-trained 200 epochs. }\vspace{-0.5cm}
\begin{center}
\tablestyle{15pt}{1.1}
\begin{adjustbox}{max width=1\textwidth}
\begin{tabular}{c|ccc|ccc|ccc}
\shline
\multirow{2}*{Model} & \multicolumn{3}{c|}{\emph{Faster R-CNN + R-50}} & \multicolumn{3}{c|}{\emph{Mask R-CNN + R-50}} & \multicolumn{3}{c}{\emph{K.P. R-CNN + R-50}} \\
~ & \apbbox{}{~} & \apbbox{50} & \apbbox{75} & \apmask{~} & \apmask{50} & \apmask{75} & \apkp{~} & \apkp{50} & \apkp{75} \\
\hline
{random} & {30.1} & {48.6} & {31.9} & {28.5} & {46.8} & {30.4} & {63.5} & {85.3} & {69.3} \\
supervised & 38.2 & 59.1 & 41.5 & 35.4 & 56.5 & 38.1 & 65.4 & 87.0 & 71.0 \\
\hline
MoCo-v1 & 37.1 & 57.4 & 40.2 & 35.1 & 55.9 & 37.7 & 65.6 & 87.1 & 71.3 \\
MoCo-v2  & 37.6 & 57.9 & 40.8 & 35.3 & 55.9 & 37.9 & 66.0 & 87.2 & 71.4 \\
\hline
SwAV &38.5&60.5&41.4&36.3&57.7&38.9&65.4&86.7&71.2\\
MoCo-M &39.3&\bf 61.1&42.5&36.8&58.4&39.4&66.1&87.3&72.4\\
LORAC  & \bf 39.5&\bf 61.1 & \bf 43.1 & \bf 37.1 & \bf 58.5 & \bf 40.2 &\bf 66.6 &\bf 87.5 &\bf 72.7\\
\shline
\end{tabular}
\end{adjustbox}
\end{center}
\label{tab:detection}
\vspace{-0.5cm}
\end{table*}

{\bf{Semi-Supervised Learning:}} We deploy semi-supervised learning task following \cite{Swav2020,chen2020simple,BYOL2020}. We finetune on classification task with either $1\%$ or $10\%$ of ImageNet1K data (same splits as \cite{chen2020simple}) on the ResNet-50 pre-trained (200 epochs) with various approaches. Specifically, we finetune the pre-trained ResNet-50 networks with a SGD optimizer, batch size of $N=256$, momentum of $0.9$ without any weight decay. We train $30$ epochs on 1\% number of ImageNet images, and train 50 epochs on the 10\% number of ImageNet images, respectively. Following SwAV \cite{Swav2020}, the learning rates of the linear layers are respectively $250 \times$ scales and $20 \times$ scales for ``1\%'' and ``10\%'' tasks, in comparison to the backbone network weights. For both ``1\%'' and ``10\%'' tasks, the optimal base learning rates of the linear layer are searched from 7 logarithomically-spaced values between $10^{-2}$ and  $10$, and are decayed by a factor of $0.2$ at the $60^{th}$ percentile and $80^{th}$ percentile of total epochs.

Table  \ref{tab:semi} reports both top-1 and top-5 accuracy on the test set. Again, we finetune a MoCo-v2 model pretrained on ImageNet1K for 200 epochs as a reference baseline, and we compare with MoCo-M, LORAC and SwAV pretrained for 200 epochs. LORAC still demonstrates clear performance boost than other approaches under consideration, especially when compared with MoCo-M and the state-of-the-art multi-view SwAV approach. This again shows LORAC's better generalization capability as an unsupervised pre-training approach, when the number of training samples from downstream tasks are limited. See \cite{Swav2020, chen2020simple} for more implementation details.

\begin{figure*}[t!]
 \centering
 \subfigure[Test accuracy vs. $\beta$.]{\label{fig:rta}
  \includegraphics[height=0.19\textwidth]{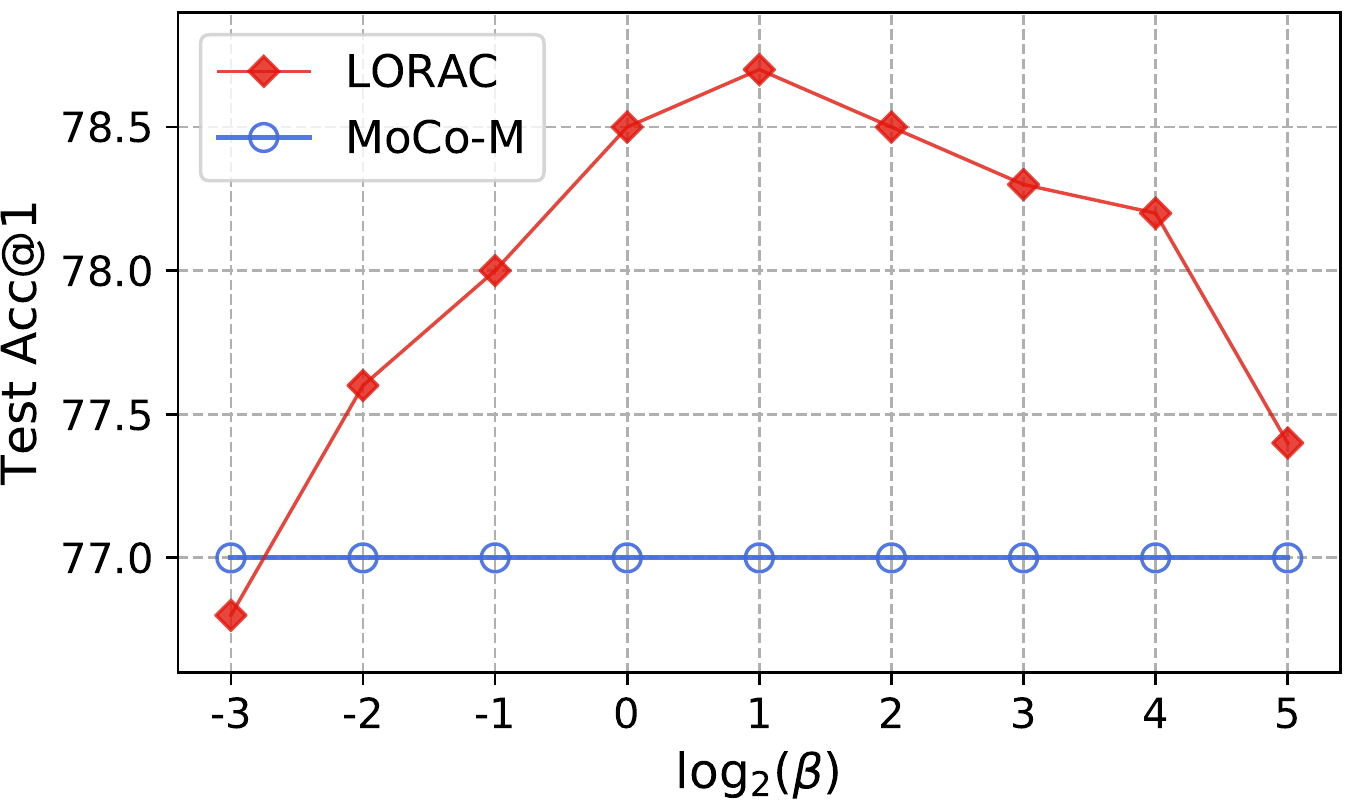}}
  \subfigure[Nuclear norm $\|\widehat \bQ\|_*$ vs. $\beta$.]{\label{fig:rtb}
  \includegraphics[height=0.19\textwidth]{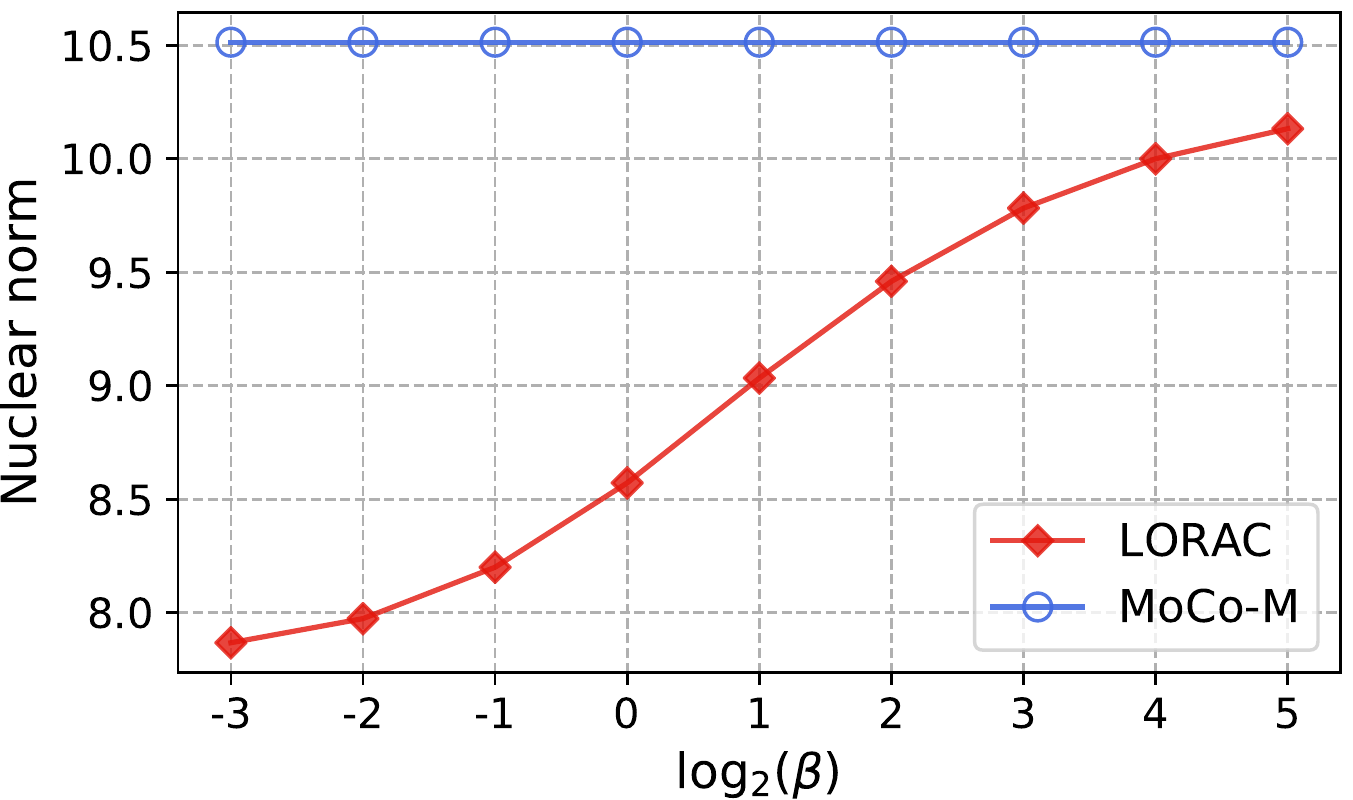}}
  \subfigure[Test accuracy vs. nuclear norm $\|\widehat \bQ\|_*$.]{\label{fig:rtc}
  \includegraphics[height=0.19\textwidth]{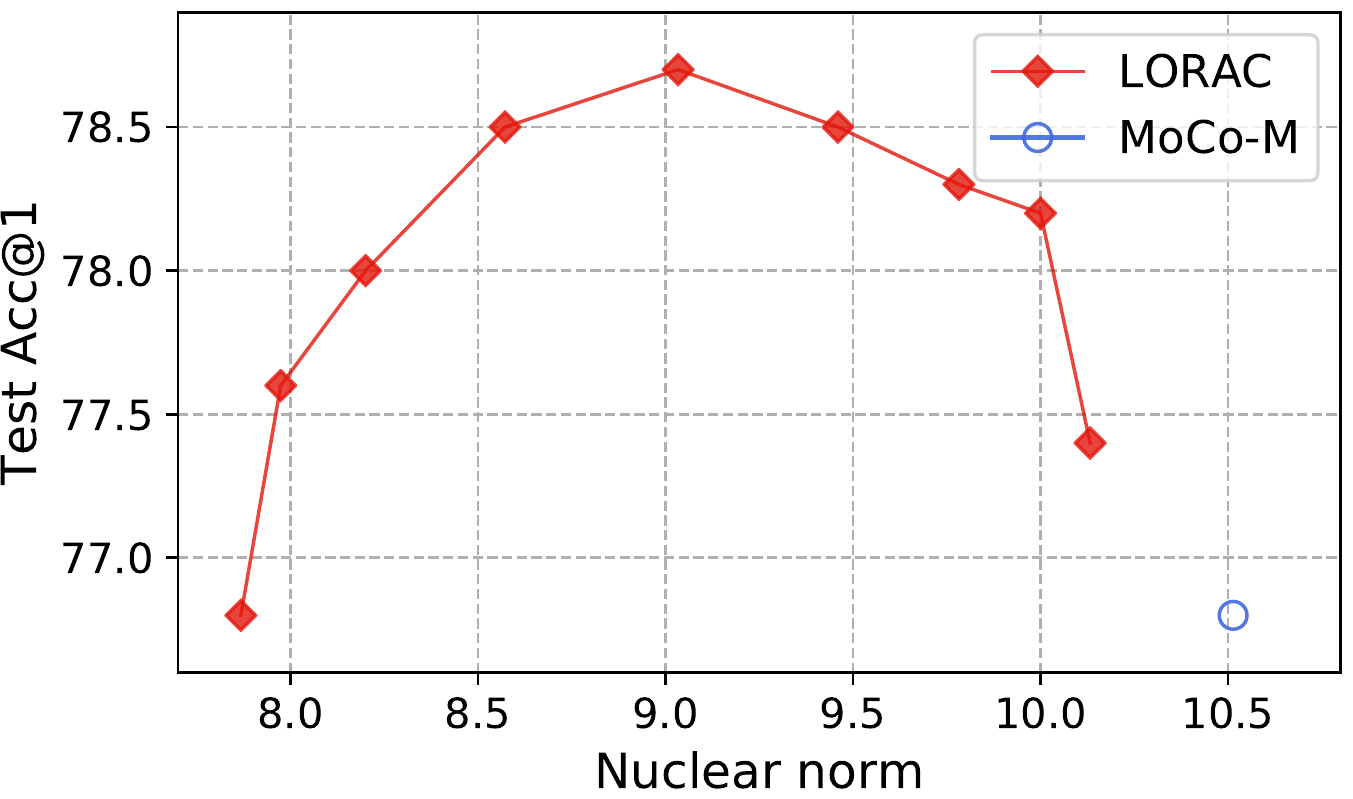}}
 \caption{\small Impact of hyperparameter $\beta$: (a) Test accuracy vs. $\beta$, (b) Nuclear norm $\|\widehat \bQ\|_*$ vs. $\beta$, (c) Test accuracy vs. nuclear norm $\|\widehat \bQ\|_*$.}
 \label{fig:rt}
\end{figure*}

{\bf Transfer to other Downstream Tasks:}
In this section, we examine LORAC's behavior on other downstream tasks ranging from instance-level to pixel-level, i.e., object detection, instance segmentation and keypoint detection. Specifically, we use the LORAC pre-trained network to initialize the backbone network ResNet-50. We then finetune the network parameters along with the attached detectors in order to examine the transfer capability of the features pre-trained by the various unsupervised algorithms.

For LORAC and MoCo-M, we finetune networks on the COCO dataset and compare the standard COCO metrics including AP (averaged over [0.5:0.95:0.05] $IoU$s), AP$_\text{50}$($IoU$=0.5) and AP$_\text{75}$($IoU$=0.75) respectively for object detection, instance segmentation and keypoint detection tasks. For object detection, we employ Faster R-CNN \cite{ren2015faster} with FPN \cite{lin2017feature} as the detector. As implemented as in \cite{he2019momentum}, we add batch normalization on the FPN layers. We perform the training on a 4-GPU machine with batch size $N=16$. We train all models for 90k iterations (${\bf 1} \times$ shedule), as in \cite{he2019momentum}.  We use the same settings for the instance segmentation and keypoint detection tasks as used for detection. The remaining hyperparameters follow the definitions in \cite{he2019momentum}.

In Table \ref{tab:detection}, LORAC outperforms other unsupervised algorithms on all of the three downstream tasks. This verifies that low rank hypothesis is legitimate and desired for improving the transfer capability of pre-trained features, even if the pre-trained network is finetuned.

\subsection{Ablation Study}\label{sec:ablation}
{\bf{Dataset and Network Setup}}:
We firstly explore the role of each predefined hyperparameters present in the LORAC loss as in Eq. (\ref{eq:Liloss}). Admittedly, it is cumbersome to present the ablation study on the entire ImageNet1K \cite{deng2009imagenet} dataset using a ResNet50  \cite{he2016deep} network owing to the huge number of training data and network parameters. Instead, following \cite{cai2020joint,chuang2020debiased,HardNegMix2020,tian2019contrastive}, we present ablation study using ResNet18 on a standard benchmark referred to as ImageNet100, a sampled subset having 100 classes from ImageNet1K. For all of the experiments, we use $\tau=0.2$. We run LORAC for 100 epochs on ImageNet100 dataset and anneal the value of $\beta$ for LORAC during training depending on the specific context. We observe similar behavior of $\tau$ as in \cite{chen2020improved} and do not include the discussion on $\tau$. For all pre-training methods, batch size is $N=128$ on 4 GPUs (32 images on each GPU).

For MoCo-v2 pretraining, we use SGD optimizer, with base learning rate 0.1, weight decay $10^{-4}$.We pre-train LORAC and MoCo-M for 100 epochs, and use MoCo-v2 (100 epochs) as a reference baseline. Data augmentation for each approach follows exactly the same protocol as used in Section \ref{sec:evllc1}. Especially, without any particular specification, the default multi-crop \cite{Swav2020} augmentation for both LORAC and MoCo-M is the 3$\times$224+5$\times$96 strategy (8 views and 10 forward strategy as defined in SwAV, and used in Section \ref{sec:evllc1}). We use LARS optimizer to train both LORAC and MoCo-M with a base learning rate $lr=2.0$ and a weight decay of $10^{-6}$. In addition, we use cosine learning rate decay schedule \cite{loshchilov2016sgdr} with a ``final learning rate'' of $0.002$ suggested by SwAV. We adopt synchronized batch normalization layers across GPUs for SwAV, as well as for the query encoder of MoCo-M and LORAC. Note that the temperature parameter $\tau$ is set to $0.1$ for SwAV while the $\tau$ value is $0.2$ for MoCo-M, LORAC by default. The reason is because the contrastive losses and SwAV loss are intrinsically different with distinct $\tau$ definitions, so we simply use the default $\tau$ values suggested by MoCo and SwAV respectively.

Throughout this section, we use top-1 classification accuracy as the primary (default) evaluation metric for comparisons, based on all model pre-trained on ImageNet100 for 100 epochs (unless particular epoch specification). We train a linear classifier on frozen features obtained from various models following the protocols in \cite{he2019momentum}. We initialize a ResNet-18 with the network parameter values (the layers till the global pooling) copied from various pre-trained models (e.g., LORAC, MoCo, MoCo-M). We then append a fully connected layer on top of this ResNet-18 backbone. During the training of downstream classification task, only the parameters of the last fully connected layer (classifier) is updated via backpropagation, while the remaining parameters of other layers are frozen. For all pre-training approaches, we train the classifier for 100 epochs on 4 GPUs. As for MoCo-v2 based linear classification, we use batch size $N=256$, SGD optimizer with $lr=30$, momentum 0.9 and without weight decay. The learning rate are respectively reduced by 0.1$\%$ respectively at 60 and 80 epoch. Regarding LORAC and MoCo-M pre-trained models, we train the linear classifier using the SGD optimizer with base learning rate 0.1 upon a cosine learning rate schedule.

\begin{figure*}[t!]
 \centering
 \subfigure[Convergence analysis]{\label{fig:convergence}
  \includegraphics[width=0.324\textwidth]{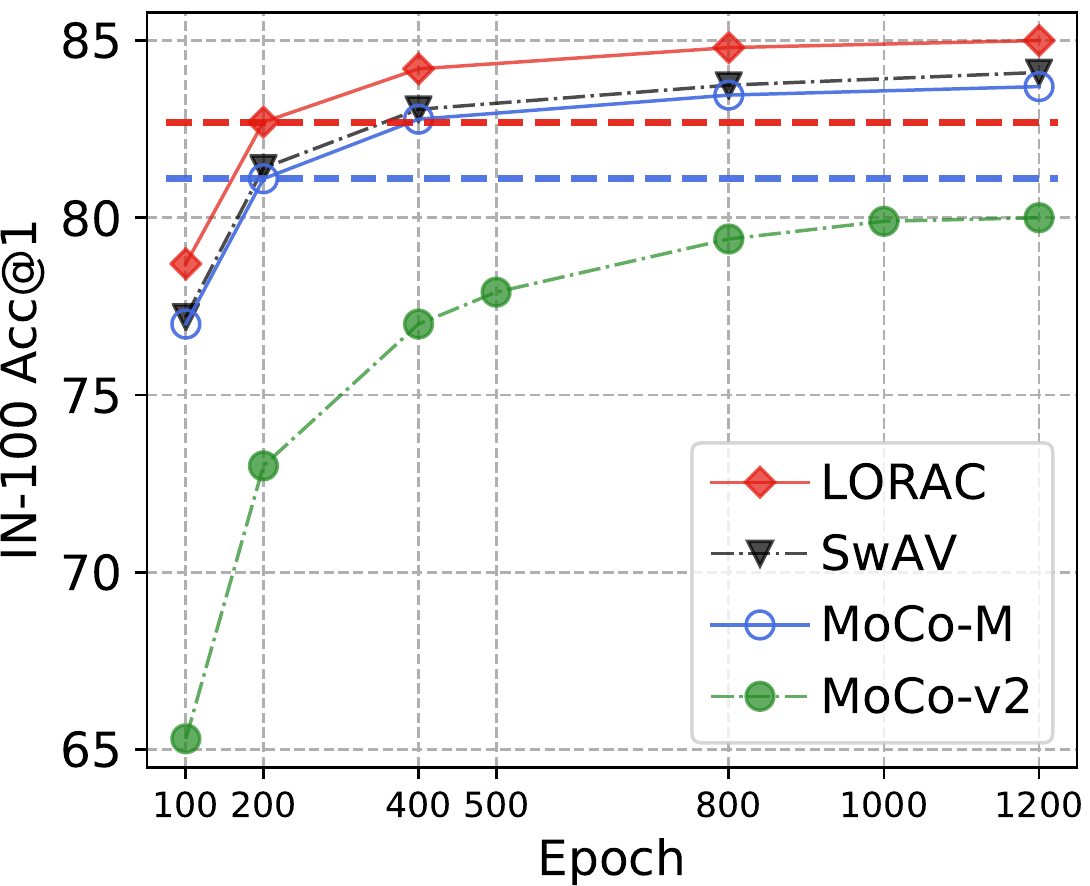}}
 \subfigure[Number of views $M$]{\label{fig:kcrop}
  \includegraphics[width=0.318\textwidth]{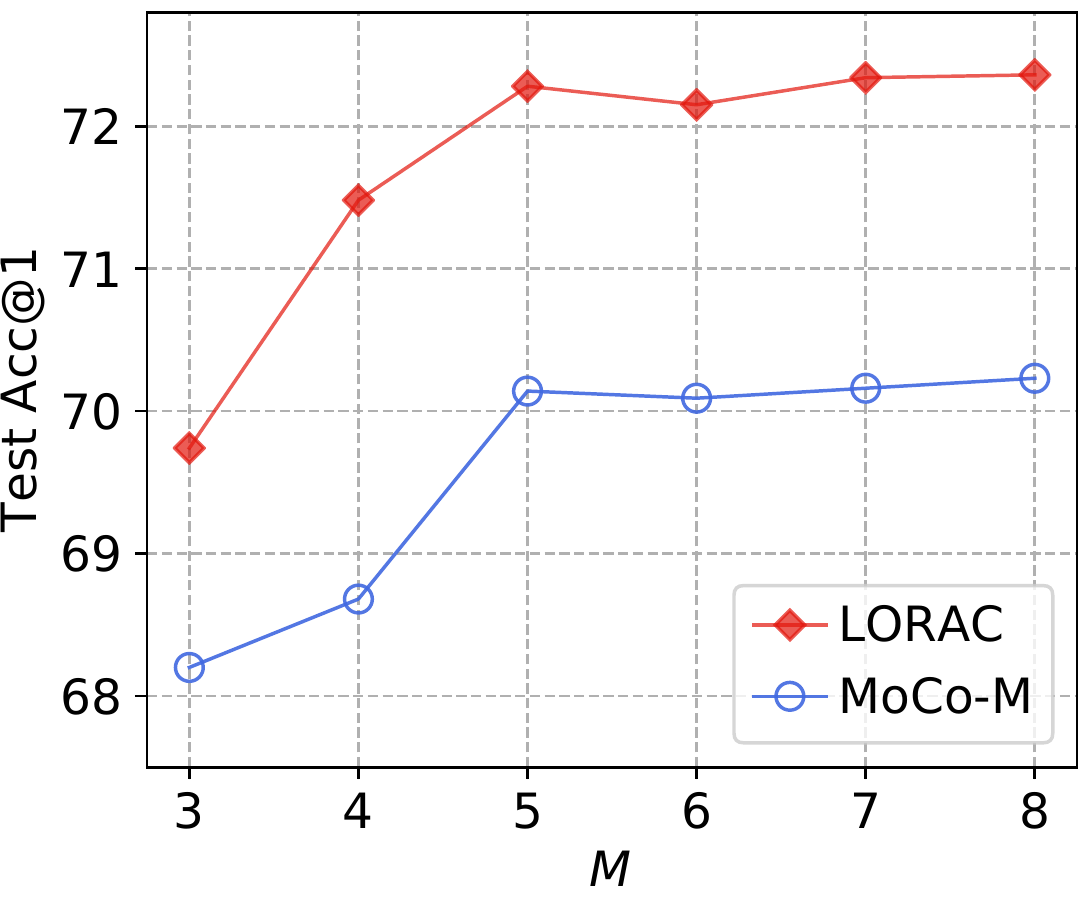}}
 \subfigure[Histogram of the nuclear norm $\widehat \bQ$]{\label{fig:nuclearnorm}
  \includegraphics[width=0.325\textwidth]{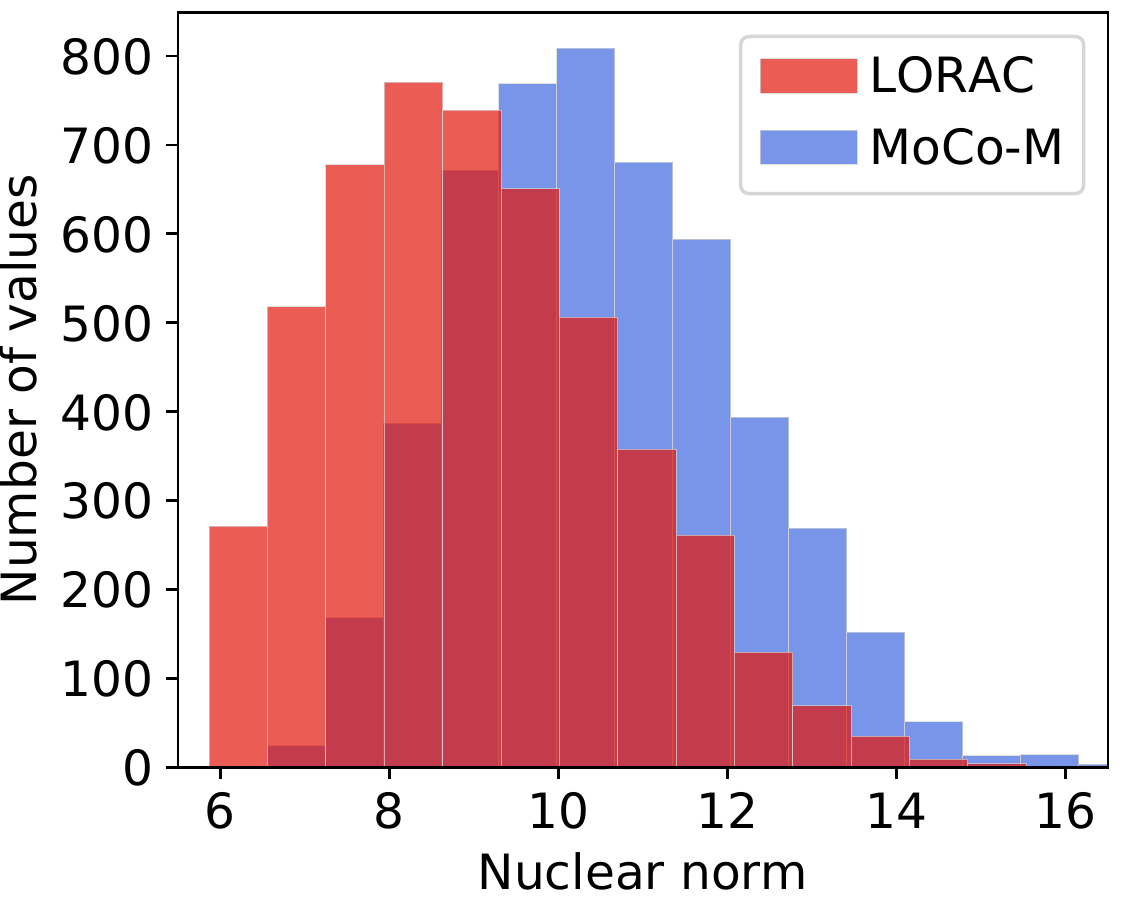}}
 \caption{\small Ablation study. (a) Convergence analysis, (b) Number of views $M$ vs. top-1 classification accuracy, (c) Histogram of the nuclear norm $\widehat \bQ$ on ImageNet100 test data.}
 \label{fig:stats}
\end{figure*}

{\bf{Impact of}} $\beta$: From the perspective of loss Eq. (\ref{eq:Liloss}), $\beta$ controls the weight of the nuclear norm and trades-off the influence of $h(\bQ)$. In principle, a relatively big $\beta$ assigns a flatter prior density on the distribution of $\|\bQ\|_*$, posing little preferences over $\bQ$ and therefore approaches the performance of MoCo-M baseline. In comparison, an extremely small $\beta$ value overly focuses on the subspace constraint while the contrastive loss w.r.t. negative pairs are significantly diluted. This risks confusing features from various instances to cluster together, thus deteriorating the performance.  We also anneal the value of $\beta$:  For each $\beta$ value illustrated in Fig. \ref{fig:rta}, we use $\beta\rightarrow \infty $ during $1-50^{th}$ epoch, the exact value of $\beta$ illustrated in Fig. \ref{fig:rta} during $51-100^{th}$ epoch. As can be seen from Fig. \ref{fig:rta}, the best top-1 classification accuracy is achieved at $\beta=2$. As $\beta$ increases, the impact of the low rank prior reduces,  gradually closing the gap between the LORAC and the MoCo-M baseline. As $\beta$ decreases to values below $1$, the prior starts to hamper the performance, owing to the diminishing impact of negative pairs in Eq. (\ref{eq:Liloss}), as per discussion in Theorem \ref{theo:1}. Actually, it can even be analytically shown that in the worst case $\beta \rightarrow 0$, all features corresponding to the entire training dataset would be mapped to the single point such that Eq. (\ref{eq:Liloss}) is reduced. In Fig. \ref{fig:rtb} and \ref{fig:rtc}, we concatenate test data features out of LORAC corresponding to 32 augmentations for each instance into a matrix $\widehat \bQ \in \mathbb{R}^{32\times128} $ (different from $\bQ\in \mathbb{R}^{M\times128}$ used for training), and compute $\|\widehat \bQ\|_*$. Fig. \ref{fig:rtb} illustrates that $\beta$ is even monotonously correlated to the change in nuclear norm on test data. There exists a sweet spot in the Fig. \ref{fig:rtc}, owing to the monotone correlation between $\beta$ and $\|\widehat \bQ\|_*$. These all reaffirm the analysis in Section \ref{sec:anl}.

{\bf{Impact of Longer Epochs (Convergence Analysis)}}: 	
This section basically verifies that LORAC is not simply accelerating the algorithms, but rather is generating distinct feature distributions against the competing approaches due to its unique advantage out of the low rank constraint. Note, we argued that low rank hypothesis has fundamentally changed the optimization landscape with reduced local minima  in comparison to independent self-supervised learning. To this point, perhaps training all the approaches till convergence on a small dataset is the best justification. This solution also, to some extent, shows that different self-supervised learning approaches converges to different local minima. Most importantly, this convergence analysis also verifies that LORAC {\textit{is neither simply benefiting from seeing more views (than MoCo-v2) or from speeding up convergence (than MoCo-M), but rather by efficiently avoiding bad local minima via optimization.}}

Figure \ref{fig:convergence} illustrates the top-1 accuracy of linear classification task pretrained and tested on ImageNet100. Note, every approach respectively reaches a performance plateau beyond 1,000 training epochs, showing convergence at different local minima. Particularly, both LORAC and SwAV have shown relatively higher accuracy than MoCo-M at the convergence, whereas the gap between MoCo-M and SwAV is very limited. A partial reason behind is the simultaneous training strategy that has effectively smoothed away bad local minima through extra grouping prior assumptions, i.e., clustering assumption for SwAV; or subspace constraint of LORAC. Notably, LORAC keeps its leading position till the eventual convergence, surpassing its multi-view counterpart SwAV, single view baseline MoCo, and most importantly, the constructed multi-view MoCo baseline MoCo-M (LORAC with $\beta \rightarrow \infty$). We believe the MoCo-M baseline has properly ablated the effect of multi-views, optimization, and augmentation policy against LORAC, although it still shows inferior performance than LORAC at the convergence. Especially, we also encourage the readers to compare between LORAC (200 epochs, dotted red), MoCo-M (200 epochs, dotted blue) and MoCo-v2 (1000 epochs), owing to a $10/2=5$ times of ``number of forward'' difference between the algorithms. One might find that LORAC (200 epochs), MoCo-M (200 epochs) are both providing much better performance than MoCo-v2 (1000 epochs). Figure \ref{fig:convergence} therefore concludes that LORAC exclusively enjoys a non-negligible benefit.

{\bf{Impact of}} $M$: To best illustrate the role of $M$ in LORAC, we adhere to the data augmentation rule defined as in MoCo-v2 ($M\times$224 crops), instead of that from SwAV multi-crop definition. This helps us remove the influence of multi-resolution in multi-crop policy and facilitate ablation only against the value of $M$ strictly defined in Eq. (\ref{eq:Liloss}). In this experiment, we use $\beta=1$. MoCo-M effectively removes the low rank prior from LORAC but still retains the usage of multiple queries. Fig. \ref{fig:kcrop} well illustrates that, LORAC improves as $M$ gradually increases and LORAC is always superior to other baselines. The curve fluctuates and almost reaches a plateau beyond $M=6$ with only slight increases. However, there remains a significant gap between LORAC and MoCo-M, which further manifests that the low rank prior itself plays a critical role in characterizing the subspace constraint for each instance.

{\bf{Impact of $h(\bQ)$} Choices}: In order to analyze the essence of hypothesis $h(\bQ)$, we further consider comparisons with the following baselines closely related to LORAC:
{\bf a)} {\em Margin based softmax}: $h(\bQ)=\exp({{- c /  (M \cdot\beta \cdot \tau) }})$, $c\in \mathbb{R}^1$ is a constant; {\bf b)} {\em Scaled Gaussian prior}: $h(\bQ)=\exp ({{- \|\bQ\|_*^2 /  (\sigma^2\cdot M \cdot \tau }}))$, $\sigma^2$ is the Gaussian variance; {\bf c)} {\em MoCo-M}, where $h(\bQ)=\exp({{-  \|\bQ\|_* /  (M\cdot\beta\cdot \tau)})}$, with $\beta\rightarrow \infty$; {\bf d)} {\em MoCo baseline} without usage of multiple views. In this section, we use view number of 8 ($3\times 224+5\times 96$). LORAC-bs stands for the LORAC loss defined in Eq. (\ref{eq:barLiloss}) with $\bP$ matrix constructed as in Section {\ref{sec:extension}} which is shared across all instances in a batch.


\begin{table}[t]
\caption{Performance comparison under various $h(\bQ)$ hypothesis pre-trained for 100 epochs on ImageNet100 dataset. Accuracy in $\%$.}
    \centering
    \tablestyle{3.6pt}{1.2}
    \begin{adjustbox}{max width=0.6\textwidth}
    \begin{tabular}[t]{c|cccccc}
        \shline
        Model  &  LORAC &   Gauss. &  LORAC-bs & L-sftmx-c &  MoCo-M & MoCo-v2  \\ \hline
 Top-1 &  78.7 & 78.8 & 78.6  & 76.8& 77.0 & 65.3 \\ \hline
 Top-5 &  94.8 &  94.9  &  94.6 & 94.2 & 94.3 & 88.3 \\ \shline
    \end{tabular}
    \end{adjustbox}
    \label{tab:epochsHq}
\end{table}

Our first constructed {\em margin based softmax} baseline (demoted as L-sftmx-c) is inspired from L-Softmax \cite{liu2016large} and its extensions. These margin based algorithms introduce an inter-class angular classification margin, in order to encourage inter-class discriminativeness and intra-class compactness. In principle, the introduced margin requires that there is a nonzero margin between class boundaries, and that the upperbound of the original softmax loss should be penalized. We argue that, LORAC does not enjoy similar merits from the introduced margin via $h(\bQ)$, in the specific context of contrastive learning. To verify, we replace the nuclear norm $\|\bQ\|_*$ in $h(\bQ)$ by a constant $c=3.0$ that has a similar value of $\|\bQ\|_*$ at training convergence. As can be seen from Table \ref{tab:epochsHq}, L-sftmx-c performs even worse than the MoCo-M baseline. This phenomenon justifies the critical role of the constructed prior involving $\|\bQ\|_*$ rather than any introduced margin. As our constructed L-sftmx-c involves multiple query features (with the modified $h(\bQ)$), it still surpasses MoCo-v2 baseline evidently.

Our method does not attempt to select the ``right'' priors. Rather, different priors can be invoked, corresponding to different hypotheses about underlying truth. Therefore, we are interested in the second related baseline: replace the scaled Laplace distribution by alternative sparse promoting prior, e.g., scaled Gaussian distribution. Specifically, we plug $h(\bQ)=\exp ({{- (\|\bQ\|_*-c_o)^2 /  (\sigma^2\cdot M \cdot \tau }}))$  into Eq. (\ref{eq:fkqQ2}), $c_o=\sqrt 5$, $\sigma^2=2$, $\tau=0.2$. Table \ref{tab:epochsHq} presents that, under Gaussian prior for $h(\bQ)$ (abbreviated as Gauss.), we yield a top-1 classification accuracy of $78.8\%$, which is comparable to the Laplace prior approach under the same background setting. This demonstrates the resilience of LORAC against various prior options, verifying our Theorem \ref{theo:1}. Our alternative low rank assumption described in Section \ref{sec:extension} , i.e., the method LORAC-bs is also showing comparable result with vanilla LORAC, corroborating the efficiency and correctiveness of our various hypothesis on subspace.

{\bf Test Data Nuclear Norm Statistics:}
We examine the unseen data statistics by LORAC to confirm generalization. We use all 5000 test images from ImageNet100 test data, and generate 32 different augmentations for each image. We input these images respectively into the LORAC, MoCo-M, and MoCo v2 pre-trained ResNet18 network to obtain features. We then concatenate features corresponding to the 32 augmentations into the each instance specific matrix $\widehat \bQ \in \mathbb{R}^{32\times128}$ (different from $\bQ \in \mathbb{R}^{M\times128}$ used for training), and compute the nuclear norm $\|\widehat \bQ\|_*$ for each image ID. Fig. \ref{fig:nuclearnorm} shows, LORAC is able to reduce the $\|\widehat \bQ\|_*$ on test data than MoCo-M. This best corroborates our hypothesis in Section (\ref{sec:anl}) that low rank promoting prior is capable of effectively encouraging higher mutual correlation even on unseen data.

\section{Conclusion}
Our proposed LORAC is the first approach to connect the low rank hypothesis with the important problem of contrastive learning. The proposed LORAC is a self-supervised learning approach both theoretically accessible and empirically useful for jointly learning invariant features. By retraining the optimization problem to lie in a constrained subspace, LORAC exclusively enjoys reduced degree of freedom of the optimization in comparison to its MoCo baselines. We show that the proposed LORAC is able to advantageously smooth away bad local minima in comparison to conventional independent contrastive learning approaches. Empirically, we pre-train a ResNet50 network on ImageNet1K and evaluate the network generalization capability on various downstream tasks. Those tasks include image classification, object detection, instance segmentation and keypoint detection. We compare with many mainstream and the state-of-the-art self-supervised learning approaches, LORAC demonstrates strong superiority over the state-of-the-arts including MoCo, SwAV, BYOL, Barlow Twins and extra. Furthermore, LORAC also generalizes well to downstream data domains that are distributed differently from the pre-training data, showing reliable generalization ability.



\ifCLASSOPTIONcaptionsoff
  \newpage
\fi

{\small
\bibliographystyle{IEEEtran}
\bibliography{egbib}
}

 \clearpage
\newpage
\appendices

\section{}

\subsection{Proof of Theorem 1}
\begin{proof}
Note that we only consider normalized feature vectors having $\ell_2$ norm equal to 1, and we define each subspace corresponding to the $i$ instance as $\mathcal{S}_i$. We firstly look at the most trivial solution where $\beta \rightarrow \infty$. In this case, we recover a special case of LORAC, i.e., MoCo-M. It is evident that the local minimum under this scenario is achieved only at $\sum_j (\bk^+_{i}-\bk^-_{i,j})\exp(\bq_{i,m}^T \bk^-_{i,j})=0$, as any distinct solutions will increase the objective.

Secondly, we consider a more complex case of LORAC where $\beta\neq \infty $ and $dim(\mathcal{S}_i)>1$. As $\|\bQ\|_*$ reduces, the singular value of the matrix $\bQ$ shrinks, and the minimum of the value is reached if all rows of $\bQ$ are identical. Since the row vectors in $\bQ$ are of unit length, such minimum nuclear norm is $\sqrt M$, $M<d$. Any other construction of $\bQ$ will necessarily increase the  $\|\bQ\|_*$ and therefore increase the loss Eq. (M.12). Correspondingly, when $\|\bQ\|_*=\sqrt M$, the $h(\bQ)$ term is minimized and vanishes, without interfering with the optimal contrastive loss value.
\end{proof}

\subsection{Additional Implementation Details of Pre-Training on ImageNet1K in Section \ref{sec:evllc1}}
In order to generate negative key features, we update the queue $\mathcal{B}$ following the proposal in \cite{he2019momentum}. In detail, we enqueue each batch of key sample $\{ \bk_i^+ \}_{i=1}^N$ sequentially into $\mathcal{B}$ while we dequeue the oldest batch of keys (depending on the batch size) in $\mathcal{B}$ during each training iteration. The augmentations used for all the experiments in the main paper are generated in the same way as in MoCo v2 \cite{chen2020improved}. Specifically, a 224$\times$224 patch is firstly randomly cropped from the resized images. Then we apply color jittering, random grayscale, Gaussian blur, and random horizontal flip to each patch. We also adopt the ShuffleBN operation as in \cite{he2019momentum} and we shuffle the data before inputting them into the network. The momentum hyperparameter \cite{he2019momentum} adopted for updating the key encoder is 0.999 and the memory size is 65,536. The optimizer employed during the pre-training is SGD with 0.9 momentum and 0.0001 weight decay.

\section{Experimental Settings for Section \ref{sec:ablation}}
In this section, we describe some supplementary implementation details of the experiments in Section \ref{sec:ablation}. The ablation study section examines baseline comparisons and hyperparameter impacts on the specific task of linear classification. In detail, we firstly pre-train the networks under various baseline models on the training data of ImageNet100 dataset. We then initialize a ResNet18 network with the parameter values copied from such pre-trained model (layers copied before the global poolings of ResNet-18). We append a fully connected layer on top of the resultant backbone and only train the classifiers on the frozen features out of the pre-trained network. All of the reported scores in the figures and tables for ablation studies are evaluated by testing the obtained classifier on the ImageNet100 test data (5,000 images in total).

{\bf Definition of $\widehat \bQ$.} The definition of $\widehat \bQ$ matrix used for producing Fig. \ref{fig:rtb} and Fig. \ref{fig:rtc} is completely distinct from the definition of $\bQ$ matrix. {\bf{\em  $\widehat \bQ$ is only constructed for illustration purpose, and was not used for training}}. To produce Fig. \ref{fig:rtb} and \ref{fig:rtc}, we use all 5,000 test images (with 5,000 image IDs) from ImageNet100 test data, and we generate 32 different augmentations for each image. Each $\widehat \bQ$ is a matrix that concatenates the features of all 32 augmentations corresponding to a certain instance, where each feature vector is extracted out of the LORAC pre-trained network. Every point illustrated in Fig. \ref{fig:rtb} and \ref{fig:rtc} exactly corresponds to the value of $\|\widehat \bQ\|_*$ averaged across 5,000 number of instances given certain contexts. Similarly, the histogram Fig. \ref{fig:nuclearnorm} illustrates such $\|\widehat \bQ\|_*$ statistics across all image IDs (5,000 points in total). In contrast, the matrix $\bQ \in \mathbb{R}^{M \times 128}$ is used for training, where $M$ is the total number of training queries $\bq_{i,m},(m\in[1,M-1])$ with additional $\bk^+_i$  used during the training procedure.

\section{More Analysis on LORAC}
As discussed in the main paper, we are motivated to discuss the essential distinctiveness between LORAC and JCL, two multi-view contrastive learning algorithms under very different hypothesis.

Firstly, and most apparently from the loss function itself, the JCL model involves an expectation over the loss and therefore {\bf necessarily} depends on specific definition of generative process such that a closed form of loss function is accessible. It is also critical for JCL to choose appropriate form of this generative distribution such that the calculation of expectation inside the $\log$ term remains tractable. Particularity, JCL uses Gaussian $ \bk_{i}^{+} \sim \mathcal{N}(\bmu_{i}, \bSigma_{i})$, where $\bk_{i}^{+}$ is positive key corresponding to the query $\bq_i$. JCL then hinges on {\bf local statistics} in each training batch to compute the $\bmu_{i}$ and $\bSigma_{i}$ in order to backpropagate the loss:
\begin{align}
	&\scriptsize {\bar{\mathcal{{L}}}}_{i,k}^{\infty}\nonumber\\
	=&\scriptsize \log \bigg[ \exp(\bq_i^T\bmu_{k_i^+}/\tau+\frac{\lambda}{2\tau^2} \bq_i^T\bSigma_{k_i^+}\bq_i)  +\sum_{j=1}^{K} \exp(\bq_i^T \bk_{i,j}^{-}/\tau)\bigg] \nonumber\\
	&\scriptsize  - \bq_i^T \bmu_{k_i^+}/\tau.
	\label{eq:eachfinalJCL}
\end{align}
In contrast, LORAC completely waives the need in computing the {\bf local statistics} during each training iteration. Although LORAC involves an SVD decomposition which seems eluding, note that an SVD decomposition for nuclear norm calculation NEVER requires the matrix to be centered via mean values. Correspondingly, LORAC is completely free from the mean and centered covariance statistics, which are even the bread and butter for JCL.

A second seemingly minor but actually essential difference is that, while JCL assumes a generative process on {\bf frozen positive keys} and empirically benefits from this assumption, LORAC heavily relies on active backpropagation on {\bf queries}, this again justifies that the mechanism behind LORAC is inherently different from JCL.

Thirdly, a seemingly much deeper inner distinctiveness is that JCL is guaranteed to reach a global minimum where all positive keys corresponding to the query shall be identical. This can be observed both from their used upper bounding technique (via Jensen's inequality where the upperbound becomes tight if and only if all positive keys are the same.), and from Eq. (\ref{eq:eachfinalJCL}) that any increase in $\Sigma_{i}> 0$ would increase the loss. However, LORAC only searches for the least number of eigenvectors that span the whole subspace including all augmentations $\bq_{i,m}, m \in [1,M-1]$ for each specific $i$ instance, where the data covairance matrix CANNOT be $0$. Otherwise, the only possibility is that all queries will be mapped to $0$ vector, which is degenerate solution.

Finally, we derive the gradient with respect to $query$ features used for backpropagation. This derivation explicitly shows that LORAC and JCL have completely different {Stationary Point}, where the gradient with respect to query vector is $0$.

\vspace{\baselineskip}
\noindent Partial derivative of \noindent{\bf JCL} loss w.r.t. $\bq_i$:
\begin{align}
&\tiny
\frac{\partial {{\mathcal{{L}}}}_i}{\partial \bq_i}=\nonumber\\
&\tiny \frac{\exp(\bq_i^T\bmu_i/\tau)\Big[\bmu_i \sum\limits_j^K \exp(\bq_i^T\bk_j^-/\tau) -\sum\limits_j^K \bk_j^- \exp(\bq_i^T\bk_j^- /\tau) -\bSigma_i \bq_i \cdot \exp(\frac{\lambda}{\tau^2} \bq_i^T\bSigma_i\bq_i)  \Big]}{ \left[ \exp(\bq_i^T\bmu/\tau+\frac{\lambda}{2\tau^2} \bq_i^T\bSigma_i\bq_i)  +\sum\limits_j^K \exp(\bq_i^T \bk_{i,j}^{-}/\tau) \right]^2 }.
\label{JCLgradient}
\end{align}

 \vspace{\baselineskip}
\noindent Partial derivative of \noindent{\bf LORAC} loss w.r.t. $\bq_{i,m}$:

\begin{align}
&\tiny
\frac{\partial {{\mathcal{{L}}}}_i}{\partial \bq_{i,m}}= \nonumber\\
&\tiny \frac{\exp(\bq_{i,m}^T\bk_i^+/\tau-|\bQ|_*/\hat\tau)\Big[ \sum\limits_j^K \exp(\bq_{i,m}^T\bk_j^-/\tau)\cdot ( \frac{ \bk_i^+}{\tau} - \frac{\bk_{i,j}^-}{\tau} -\frac{\partial |\bQ|_*}{\hat \tau \partial \bq_{i,m}} )\Big]}{ \left[ \exp(\bq_{i,m}^T\bk_i^+/\tau  - |\bQ|_*/\hat\tau )  +\sum\limits_j^K \exp(\bq_{i,m}^T \bk_{i,j}^{-}/\tau)   \right]^2 }.
\label{eq:LORACgradient}
\end{align}
where $\hat\tau={M\cdot \beta\cdot \tau\cdot}$ takes into account the effect of hyperparameters as defined in Eq. (M.11). To facilitate the computation, we define the SVD decomposition of $\bQ$ as:
\begin{align}
\small
\bQ=\bU\Sigma\bV^T,
\end{align}
where $\bSigma$ is a rectangular diagonal matrix (entirely different definition from $\bSigma_i$ in Eq. (S.\ref{JCLgradient})), $\bU$ and $\bV^T$ are real orthogonal matrices. Note that:
\begin{align}
\small
\frac{\partial |\bQ|_*}{\partial \bQ}=\bU\bV^T,
\label{eq:gradient}
\end{align}

Plugging back the Eq. (S.\ref{eq:gradient}) into Eq. (S.\ref{eq:LORACgradient}) and apply chain rule, this concludes that Eq. (S.\ref{eq:LORACgradient}) has a stationary point completely different than that of Eq. (S.\ref{JCLgradient}). Note in Eq. (S. \ref{eq:LORACgradient}), we see the optimization direction is affected by the eigenvector $\bU$ that span the subspace of $\bQ$ regardless of singular value itself, whereas in JCL, the partial derivative is influenced by {\bf centered} positive key variance matrix, two irrelevant things.

It is now crystal clear that JCL and LORAC are two fundamentally different algorithms, having distinct motivation, loss function, converge at different stationary point, and different performance as well.



\end{document}